%% file: main.tex
\documentclass[lettersize,journal]{IEEEtran}
\usepackage{pifont}
\input{math_commands.tex}

\ifCLASSOPTIONcompsoc
\usepackage[nocompress]{cite}
\else
\usepackage{cite}
\fi
\ifCLASSINFOpdf
\usepackage[pdftex]{graphicx}
\usepackage{multirow}
\usepackage{amsmath}
\usepackage{amssymb}
\usepackage{booktabs}

\usepackage{xcolor} 
\usepackage{color} 
\usepackage{verbatim} 
\usepackage{diagbox}
\usepackage[pagebackref=true,breaklinks=true,colorlinks,citecolor=blue,linkcolor=blue,bookmarks=false]{hyperref}
\else

\fi
\usepackage{graphicx}  
\usepackage{float}
\usepackage{subfigure}
\usepackage{makecell}

\usepackage{colortbl}
\definecolor{yyellow}{rgb}{1, 1, 0.7}
\definecolor{oorange}{rgb}{1, 0.85, 0.7}
\definecolor{rred}{rgb}{1, 0.7, 0.7}

\usepackage{ragged2e}
\usepackage{caption}
\newcommand{\dui}{\ding{51}}
\newcommand\chn[1]{}

\newif\ifhighlight
\highlightfalse
\ifhighlight
  \newcommand{\revise}[1]{{\color{red}{#1}}} 
\else
  \newcommand{\revise}[1]{#1} 
\fi

\begin{document}
\title{AniMer+: Unified Pose and Shape Estimation Across Mammalia and Aves via Family-Aware Transformer}

\author{Liang An\(^*\), \revise{\IEEEmembership{Member, IEEE}}, Jin Lyu\(^*\), 
        Li Lin, 
        Pujin Cheng,
        Yebin Liu, \IEEEmembership{Member, IEEE}, \\
        and Xiaoying Tang, \IEEEmembership{Senior Member, IEEE}
\thanks{\(^*\) indicates equal contribution, \revise{in an alphabetical order}.}
\thanks{\revise{Liang An} and Jin Lyu are with Department of Electronic and Electrical Engineering, Southern University of Science and Technology, Shenzhen, China.}
\thanks{Liang An and Yebin Liu are with Department of Automation, Tsinghua University, Beijing, China.}
\thanks{Li Lin and Pujin Cheng are with Department of Electronic and Electrical Engineering, Southern University of Science and Technology, Shenzhen, China, also with Jiaxing Research Institute, Southern University of Science and Technology, Jiaxing, China, and also with Department of Electrical and Electronic Engineering, the University of Hong Kong, Hong Kong, China.}
\thanks{Xiaoying Tang is with Department of Electronic and Electrical Engineering, Southern University of Science and Technology, Shenzhen, China, and also with Jiaxing Research Institute, Southern University of Science and Technology, Jiaxing, China.}
\thanks{Corresponding author: Xiaoying Tang (tangxy@sustech.edu.cn).}
}


\IEEEtitleabstractindextext{
\input{src/00_abstract}
}
\maketitle
\IEEEdisplaynontitleabstractindextext
\IEEEpeerreviewmaketitle

\input{src/01_intro}

\input{src/02_related}
\input{src/03_method}
\input{src/04_experiment}

\input{src/05_conclusion}

\bibliographystyle{unsrt}
\bibliography{main}

\end{document}

%% file: math_commands.tex
\usepackage{amsmath,amsfonts,bm}

\def\eqref#1{equation~\ref{#1}}

\def\1{\bm{1}}

\DeclareMathAlphabet{\mathsfit}{\encodingdefault}{\sfdefault}{m}{sl}
\SetMathAlphabet{\mathsfit}{bold}{\encodingdefault}{\sfdefault}{bx}{n}



%% file: src/00_abstract.tex
\begin{abstract}

In the era of foundation models, achieving a unified understanding of different dynamic objects through a single network has the potential to empower stronger spatial intelligence.
Moreover, accurate estimation of animal pose and shape across diverse species is essential for quantitative analysis in biological research. However, this topic remains underexplored due to the limited network capacity of previous methods and the scarcity of comprehensive multi-species datasets. To address these limitations, we introduce AniMer+, an extended version of our scalable AniMer framework. In this paper, we focus on a unified approach for reconstructing mammals (mammalia) and birds (aves). A key innovation of AniMer+ is its high-capacity, family-aware Vision Transformer (ViT) incorporating a Mixture-of-Experts (MoE) design. Its architecture partitions network layers into taxa-specific components (for mammalia and aves) and taxa-shared components, enabling efficient learning of both distinct and common anatomical features within a single model. 
To overcome the critical shortage of 3D training data, especially for birds, we introduce a diffusion-based conditional image generation pipeline. This pipeline produces two large-scale synthetic datasets: CtrlAni3D for quadrupeds (about 10k images with pixel-aligned SMAL labels) and CtrlAVES3D (about 7k images with pixel-aligned AVES labels). To note, CtrlAVES3D is the first large-scale, 3D-annotated dataset for birds, which is crucial for resolving single-view depth ambiguities. Trained on an aggregated collection of 41.3k mammalian and 12.4k avian images (combining real and synthetic data), our method demonstrates superior performance over existing approaches across a wide range of benchmarks, including the challenging out-of-domain Animal Kingdom dataset. Ablation studies confirm the effectiveness of both our novel network architecture and the generated synthetic datasets in enhancing real-world application performance. The project page with code and data can be found at \url{https://animerplus.github.io/}. 

\end{abstract}

%% file: src/01_intro.tex
\section{Introduction}\label{sec:introduction}
\IEEEPARstart{A}{ccurate} estimation of animal pose and shape from images is essential for capturing animal behavior, biomechanics, and interactions with environment. 
It provides vital insights for diverse fields including animal welfare, agriculture, ecology, and life sciences. Integrating geometric and appearance information from widely varying species, such as birds and mammals, into a unified deep neural network represents a crucial step towards comprehensively interpreting the intricate and dynamic animal world.
Significant efforts in this domain involve estimating pose and shape parameters using articulated animal templates. For example, the well-known SMAL model~\cite{zuffi20173d} primarily targets mammals, while the AVES parametric model~\cite{wang2021birds} focuses on birds. Although extensive research has been conducted on specific species or families, such as horses~\cite{zuffi2024varen,li2021hsmal,li2024dessie}, zebras~\cite{zuffi2019three}, dogs~\cite{Sabathier2024AnimalAR,ruegg2023bite,rueegg2022barc,li2021coarse,biggs2020wldo,biggs2019creatures} and birds~\cite{wang2021birds}, investigations into other mammal species, including cats, cows or hippos, remain relatively underexplored, and few previous methods try to predict the meshes for animals of different classes  (e.g., mammalia and aves) in a single network. 

Reconstructing multiple species within a single network is hindered by two primary challenges: the limited capacity of backbones and the scarcity of multi-species datasets annotated with corresponding parametric model labels. Recent advancements in human mesh recovery via the SMPL model~\cite{SMPL:2015} demonstrate that combining high-capacity Transformer~\cite{vaswani2017attention} backbones with large-scale datasets significantly enhances pose and shape estimation accuracy across diverse settings~\cite{goel2023humans,pavlakos2024reconstructing,cai2024smpler}. However, this effective paradigm has yet to be fully explored within animal studies. While ~\cite{xu2023animal3d} introduces the large-scale Animal3D dataset featuring SMAL annotations, their work primarily utilizes CNN-based networks like HMR~\cite{kanazawa2018end} and WLDO~\cite{biggs2020wldo}.

\begin{figure*}[htbp]
    \centering
    \includegraphics[width=1.0\linewidth]{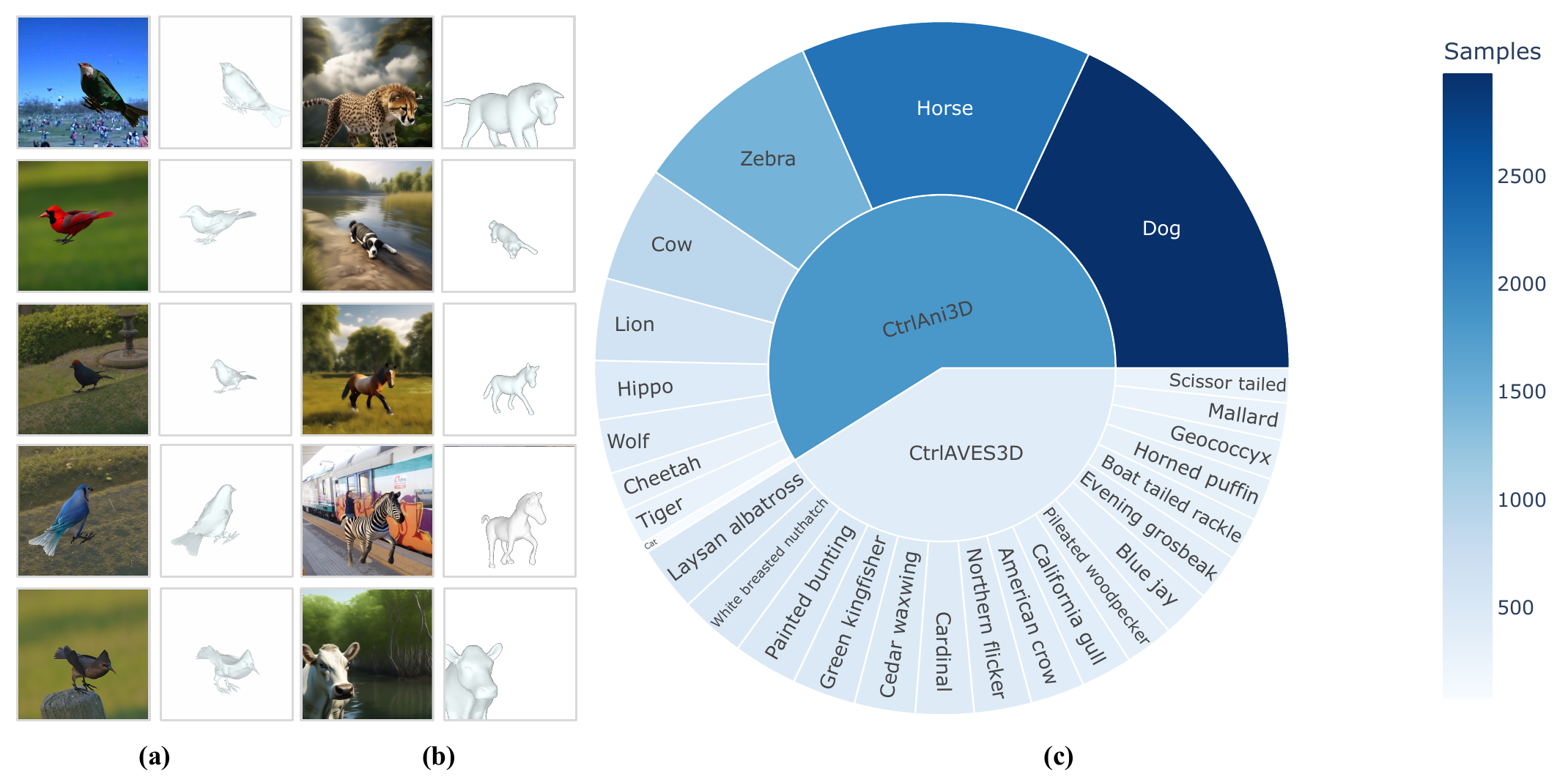}
    \caption{\textbf{Statistics and representative samples of the CtrlAVES3D and CtrlAni3D datasets.} 
    \textbf{(a)} and \textbf{(b)} represent samples from the CtrlAVES3D dataset and the CtrlAni3D dataset, respectively. For each image pair, the left side displays the generated animal image whose background comes from either COCO ~\cite{lin2014microsoft} or AI-synthesis, and the right side presents the rendered mesh label. Note that the synthesis process naturally considers generating truncated images. \textbf{(c)} shows the statistics of both datasets. More details about these two datasets can be found in Table~\ref{tab:taxonomy}, Table~\ref{tab:taxonomy_bird} and Sec.~\ref{sec:ctrlani3d}.
    }
    \label{fig:dataset_samples}
\end{figure*}


To recap, our previous work, AniMer~\cite{lyu2025animer},  proposes a systematic approach that pursues accurate \textit{\textbf{ani}}mal pose and shape estimation by leveraging family-aware Transfor\textit{\textbf{mer}}. The success of AniMer is built on two key scaling factors: a scaled backbone and a scaled dataset. Architecturally, AniMer integrates a high-capacity Vision Transformer (ViT) backbone with a Transformer-based decoder to regress SMAL parameters, a paradigm proven effective in human mesh recovery but previously untested for animal shape and pose estimation. To unify the discriminative understanding of diverse animal shapes within a single framework, AniMer introduces a novel animal family supervised contrastive learning scheme. It employs a learnable class token to explicitly enhance the model's ability to distinguish between the unique shapes of different mammal families.

A high-capacity model's effectiveness is intrinsically linked to the quality and quantity of its training data. However, acquiring a large number of animal images with full 3D annotations is difficult, in which situations synthetic datasets can mitigate the scarcity of real data. The coarse texture quality and sophisticated lighting and shadow control hinder the usability of traditional Computer Graphics (CG) generated animal images. 
These limitations motivate our usage of ControlNet~\cite{zhang2023adding} to create high-quality AI hallucinated images conditioned on SMAL structures. 
Compared to CG-based, this principle effectively bridges the domain gap between parametric labels and images, and helps to create a novel, large-scale dataset named CtrlAni3D with minimal human labor involved. 
Specifically, we prompt ControlNet with textual descriptions of animal behaviors and rendered SMAL mask and depth images, resulting in highly realistic visual outputs, as illustrated in Fig.~\ref{fig:dataset_samples}. To ensure dataset quality, we use SAM2.0~\cite{ravi2024sam2} together with manual verification to filter inappropriate images. Finally, CtrlAni3D comprises 9711 images annotated with pixel-aligned SMAL meshes, and its scalable nature allows for further expansion. We end up with scaled training datasets by aggregating most available open-sourced quadrupedal datasets and our CtrlAni3D dataset, resulting in a comprehensive set of 41.3k images annotated with either 3D meshes or 2D keypoints.

Despite these advances, the SMAL model is insufficient for animals like birds, whose skeletal structures fundamentally differ from mammals'. Training a shared model naively on both mammals and birds risks ``averaging out" crucial anatomical features and suffers from severe depth ambiguities in single-view reconstruction, exacerbated by the fact that existing avian datasets provide only 2D annotations. Inspired by the success of Mixture-of-Experts (MoE) architectures~\cite{shazeer2017outrageously,jiang2024mixtral,xu2023vitpose++} in handling diverse tasks, we propose \textbf{AniMer+}, an extension of AniMer incorporating an MoE design to better unify different animal taxa. We adapt the ViT encoder by partitioning its fully-connected (FC) layers into taxa-specific layers (capturing expert quadruped/avian knowledge) and a taxa-shared layer (learning common semantic features). The taxa-specific layers process features from the corresponding taxa, while the taxa-shared layer receives features from all taxa. This architecture enables effective handling of both taxonomic groups within a single, unified framework. To enable robust 3D training for birds, we apply our generation pipeline to create CtrlAVES3D, the first large-scale avian dataset with 3D AVES model annotations, thereby mitigating the single-view ambiguity issue. 

To rigorously evaluate the efficacy of our AniMer and AniMer+ models, as well as our dataset generation pipeline, we conduct extensive empirical studies. For a clarification purpose, we name AniMer trained on mammals only, birds only and both mammals and birds respectively as AniMer-M, AniMer-A and AniMer-AM. Our findings indicate that AniMer, when trained on the complete multi-species mammal datasets (AniMer-M), significantly outperforms CNN-based methods such as HMR and WLDO~\cite{biggs2020wldo} which serve as baselines in the Animal3D research. Similarly, AniMer-A largely surpasses previous methods on bird datasets. Meanwhile, through ablation studies, we verify that the synthesized datasets enhance the generalizability of AniMer on the out-of-domain (OOD) Animal Kingdom dataset~\cite{Ng_2022_CVPR} and the OOD Cow Bird dataset~\cite{badger20203d}, which are unseen during training. In addition, ablation studies also prove that the animal family-supervised contrastive learning scheme improves pose and shape estimation precision. Finally, the joint training of AniMer+ on both mammals and birds outperforms AniMer-AM, demonstrating the effectiveness of the MoE design.


Our contributions can be summarized as follows:
\begin{itemize}
\item 
We propose AniMer, an animal pose and shape estimation network benefiting from a high-capacity Transformer backbone and an integrated dataset covering most available mammal datasets. AniMer achieves state-of-the-art results on both in-domain datasets and OOD datasets, for both mammals and birds.  
\item 
We propose an animal family supervised contrastive learning scheme, which enhances the
model’s ability to distinguish between the unique shapes of different mammal/bird families. Experiments indicate that this scheme improves the overall performance across species, especially for animals with limited training samples.  
\item 
We propose a synthetic dataset generation pipeline based on ControlNet, and generate the CtrlAni3D dataset for mammalia and the CtrlAVES3D dataset for aves. To note, CtrlAVES3D is the first large-scale avian dataset with 3D annotations, mitigating the depth ambiguity issue the single-view reconstruction tasks. 
\item 
We further present AniMer+, the first method to unify the prediction of distinct animal taxa such as mammalia and aves. With a specialized MoE-based architecture, AniMer+ outperforms AniMer when trained on mammals and birds simultaneously, pioneering a way for future foundational animal mesh recovery network. 
\end{itemize}

An earlier version of this work is AniMer~\cite{lyu2025animer}, which was published in CVPR 2025. In this paper, we have significantly enhanced AniMer in three aspects.
First, we create the first large-scale bird dataset with 3D annotations, named CtrlAVES3D, which fulfills the lack of 3D annotations in existing avian datasets and mitigates depth ambiguity in the monocular bird reconstruction task.
Second, we propose AniMer+ by incorporating an MoE design to better unify different animal taxa, which outperforms AniMer when trained simultaneously on mammals and birds. 
Third, we conduct comparative and ablation experiments on avian species, establishing state-of-the-art performance across multiple avian datasets.
Based on these extensions and in conjunction with the dataset generation pipeline, AniMer+ has the potential to generalize to any animal groups (e.g., fishes, insects, rodents or primates) that can be represented by parametric models.

%% file: src/02_related.tex
\section{Related Works}

\noindent\textbf{Pose and Shape Estimation for Mammals and Birds.}
Compared to monocular 2D/3D animal keypoint estimation methods~\cite{mathis2018deeplabcut,bala2020automated,pereira2022sleap,ye2024superanimal,dabhi20243d}, monocular surface estimation methods can provide both shape and pose information simultaneously.
Previously, template-free methods can reconstruct mammals~\cite{yang2022banmo,yao2022lassie,jakab2024farm3d,li2024learning,kaye2025dualpm} or birds~\cite{kanazawa2018learning} with arbitrary topology. Although these methods demonstrate high flexibility, their surface accuracy still has room for improvement. 
In this paper, we focus on template-based methods instead of template-free ones. 
Reconstructing shape templates has been studied in various types of animal families such as birds~\cite{badger20203d, wang2021birds}, rodents~\cite{an2023three}, non-human primates~\cite{neverova2020continuous} and quadrupeds~\cite{zuffi2018lions}. For quadrupeds, the well-known SMAL~\cite{zuffi20173d} is built upon 41 scans of animal toys. Due to the limited geometry accuracy of SMAL for representing specific species, most previous methods stride over predicting SMAL parameters only and further enhance the geometry of horses and dogs. For example, ~\cite{li2021hsmal} designs the horse-specific hSMAL model and applies it to the video lameness detection problem. VAREN~\cite{zuffi2024varen} further improves hSMAL with high quality horse scans. Similarly, the use of SMAL for dog mesh recovery is addressed by adding bone lengths control~\cite{biggs2020wldo}, by adding per-vertex deformation~\cite{li2021coarse}, by introducing a breed loss~\cite{rueegg2022barc}, by ground contact constraints~\cite{ruegg2023bite}, or by temporal avatar optimization~\cite{Sabathier2024AnimalAR,zhong20254d}. Though high quality reconstruction has been achieved on horses and dogs, end-to-end SMAL estimation has not been fully addressed on other challenging species. Animal3D~\cite{xu2023animal3d} provides the first large-scale 3D benchmark for general quadruped SMAL estimation, yet neglects in-depth research on network design and training strategy.  

\noindent\textbf{Transformer Based Human Mesh Recovery.}
The Transformer architecture~\cite{vaswani2017attention} has revolutionized the field of Natural Language Processing (NLP) by enabling unprecedented accuracy and efficiency in a wide range of tasks. 
Inspired by its success in NLP, one core part of Transformer, i.e. self-attention, has been widely employed in human mesh recovery~\cite{kocabas2021pare, wan2021encoder, shen2023global, shin2024wham}. ~\cite{dosovitskiy2020image} proposes ViT, which divides an image into patches as the input to Transformer. ViT has achieved state-of-the-art performance on several computer vision tasks including human mesh recovery using SMPL~\cite{kocabas2021pare, wan2021encoder, shen2023global, goel2023humans, shin2024wham}. Among all these previous works, HMR2.0~\cite{goel2023humans} is a milestone which demonstrates the effectiveness of simply using a ViT backbone and large-scale datasets to achieve accurate mesh recovery and in-the-wild generalizability. Inspired by HMR2.0, HaMeR~\cite{pavlakos2024reconstructing} employs a ViT backbone to achieve highly accurate hand mesh recovery, which is further extended to interacting hands~\cite{lin20244dhands}. Similarly, SMPLer-X~\cite{cai2024smpler} scales up expressive human pose and shape estimation using a ViT backbone and a combination of 32 datasets. Although impressive results have been achieved in human mesh recovery, the effect of employing a ViT backbone for animal pose and shape estimation remains unexplored. 


\noindent\textbf{Synthetic Animal Training.}
Compared to human pose estimation which benefits from large-scale datasets, acquiring annotated images of animals is significantly more difficult. Synthetic datasets shall alleviate this issue by rendering the input and output simultaneously. Most previous methods only attempt to render RGB images~\cite{cao2019cross,mu2020learning,plum2023replicant,ruegg2023bite,shooter2024digidogs,xu2023animal3d,zhong2025moremouse} or depth images~\cite{Kearney_2020_CVPR} using traditional CG pipelines, ignoring the image hallucination ability of generative AI models such as stable diffusion~\cite{rombach2022high} or ControlNet~\cite{zhang2023adding}. 
Although traditional CG-based rendering achieves success in assisting network training, large-scale and high-quality CG assets are expensive to obtain, and the usability of rendered images is hindered by sophisticated lighting and shadow control. In contrast, ~\cite{ma2023generating} uses 3D
visual prompts and large language model (LLM) text prompts to generate diverse images from a static CAD model, pioneering a new direction. We extend their idea to animal image synthesis from dynamic SMAL animations to construct the CtrlAni3D dataset. \cite{jiang2023spac} performs style-transfer from synthetic domain to real domain using ControlNet~\cite{zhang2023adding}. However, they rely on textured CAD assets instead of untextured SMAL, making it difficult to assist SMAL pose and shape estimation. Concurrent work GenZoo~\cite{niewiadomski2024generative} generates realistic images with corresponding SMAL+~\cite{zuffi_eccv2024_awol} parameters utilizing ControlNet in a similar way as ours, but they are specific to quadrupeds. 

%% file: src/03_method.tex

\begin{figure*}[htbp]
    \centering
    \includegraphics[width=\linewidth]{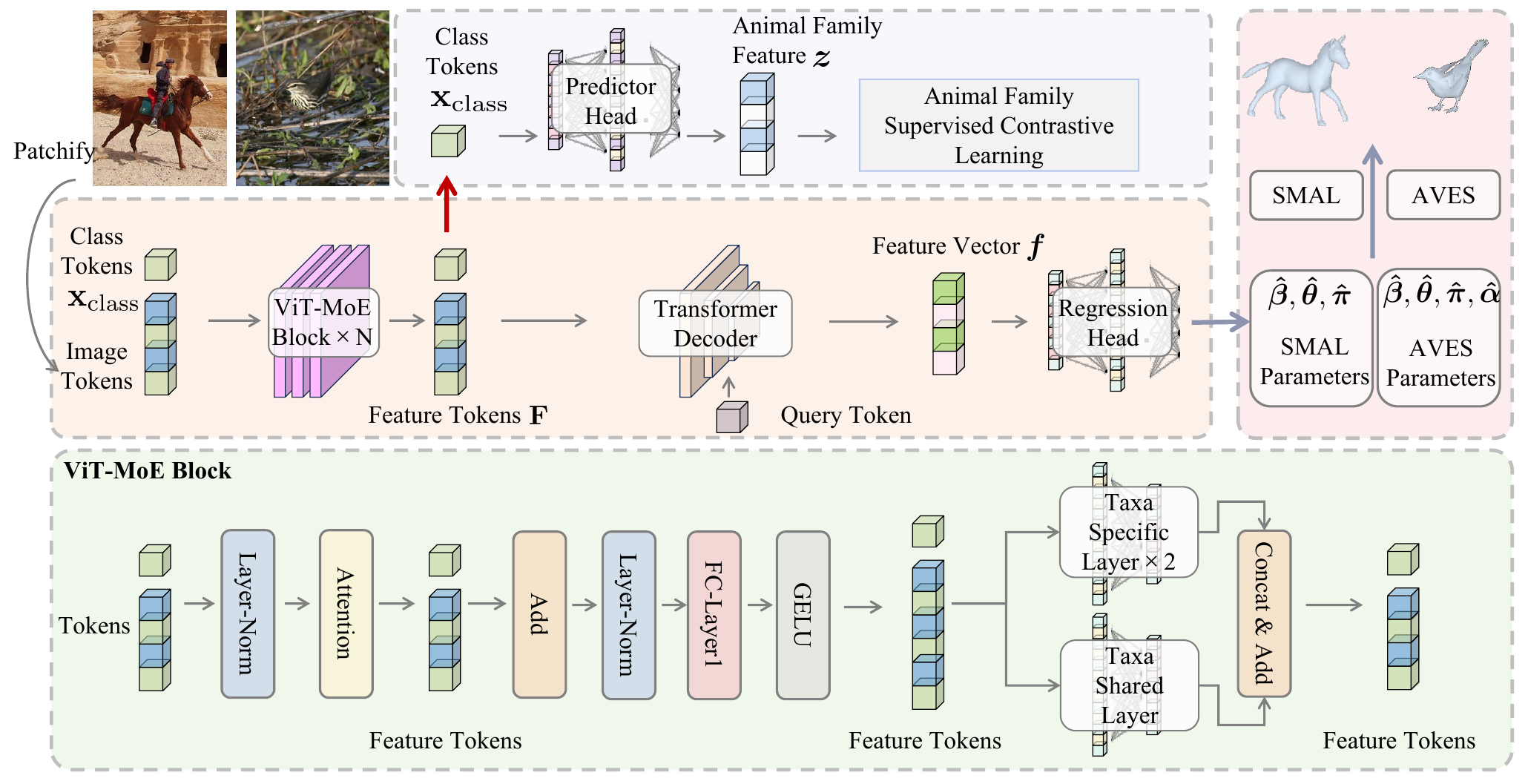}
    \caption{\textbf{AniMer+ network architecture.} AniMer+ consists of (1) a ViT-MoE encoder with N ViT-MoE blocks that extract quadrupedal or avian image features; (2) a Transformer decoder that processes features generated by the encoder; (3) a predictor head (MLPs) that generates animal family features for supervised contrastive learning; and (4) a regression head (MLPs) that estimates the parametric model parameters.}
    \label{fig:animerplus_pipeline}
\end{figure*}

\section{Method}
\subsection{Preliminaries}
\noindent\textbf{The SMAL Model.} The SMAL model~\cite{zuffi20173d}, denoted as \(\mathcal{M}(\beta, \theta, \gamma)\), is a 3D parametric shape model designed for quadrupedal mammals. The shape parameter \(\beta\ \in \mathbb{R}^{41} \) is derived from 41 3D scans of various animal figurines, such as cats, dogs, horses, cows, and hippos. The pose parameter \(\theta \in \mathbb{R}^{35 \times 3}\) represents the rotation of each joint relative to its parent joint, expressed in terms of axis-angle. Controlled by \(\beta\) and \(\theta\), the SMAL model outputs a mesh with vertices \(V \in \mathbb{R}^{3889 \times 3}\) and faces \(F \in \mathbb{N}^{7774 \times 3}\) through a linear blend skinning (LBS) process. The animal body joints are regressed from vertices by \(J \in \mathbb{R}^{35 \times 3} = W \cdot V\), where $W\in\mathbb{R}^{35\times3889}$ represents a linear mapping matrix.

\noindent\textbf{The AVES Model.} AVES is a multi-species avian model that learns a PCA shape space on 17 avian species. AVES can be represented by a function \(\mathcal{M}(\beta, \theta, \alpha, \gamma)\), where \(\beta \in \mathbb{R}^{15}, \theta \in \mathbb{R}^{25 \times 3}\) and \(\gamma \in \mathbb{R}^{3}\). Unlike SMAL, AVES adds a bone parameter \(\alpha \in R^{24}\) to scale the distance between neighboring joints. Given \(\beta, \theta\) and \(\alpha\), the AVES model returns a triangle mesh with vertices \(V \in \mathbb{R}^{8210 \times 3}\), faces \(F \in \mathbb{N}^{12468 \times 3}\) and joints \(J \in \mathbb{R}^{25 \times 3}\).

\noindent\textbf{Camera Projection.} Following HMR2.0~\cite{goel2023humans}, \(\pi(\cdot)\) represents the projection process of a weak-perspective camera model, determined by a translation vector $T\in\mathbb{R}^3$, a fixed focal length and thereby a fixed intrinsic matrix $K$. The global orientation $R$ is the rotation of SMAL's root joint, therefore we ignore it here. Consequently, a 3D point $X$ is projected as a 2D point $x$ by $x=\pi(X)=\Pi(K(X+T))$, where $\Pi$ converts homogeneous coordinates $(u,v,d)^T$ to pixel coordinates $(u/d,v/d)^T$. 

\subsection{The Architecture of AniMer}
Given an image \(I \in \mathbb{R}^{H \times W \times 3}\) and a class token \(\mathbf{x}_{\text{class}} \in \mathbb{R}^{1 \times 1280}\), we first utilize a ViT encoder to extract image feature tokens \(\mathbf{F} \in \mathbb{R}^{192 \times 1280}\), and the class token interacts with the image to capture information about the animal family. We then feed the feature tokens \(\mathbf{F}\) into a SMAL Transformer decoder to obtain a feature vector \(\boldsymbol{f} \in \mathbb{R}^{1 \times 1024}\). Finally, independent multi-layer perceptrons (MLPs) are used to predict the shape parameters \(\hat{\beta}\), pose parameters \(\hat{\theta}\), and camera parameters \(\hat{\pi}\), where $\hat{\cdot}$ means estimated parameters. Meanwhile, the class token is fed into the predictor head for animal family supervised contrastive learning, which will be detailed in Sec.~\ref{sec:sec:const}. Note that we borrow the pretrained weights of the ViT encoder from ViTPose++~\cite{xu2023vitpose++}, which significantly enhances mesh recovery performance.
 
Despite the class token, AniMer features two key differences compared with previous HMR2.0 and HaMeR~\cite{pavlakos2024reconstructing}. 
First, both HMR2.0 and HaMeR regress the residual SMPL/MANO parameters with respect to the non-zero mean parameters computed from large-scale motion databases. In contrast, AniMer directly decodes the final SMAL parameters due to limited SMAL pose prior. Second, both HMR2.0 and HaMeR train on the whole datasets in a single stage. Instead, we train AniMer in two stages. 
At the first stage, we train AniMer using only 3D data to ensure the network can predict feasible shapes and poses. At the second stage, all 3D and 2D data are introduced for training. 
The insight is that the size of 3D animal datasets is much smaller than that of 2D animal datasets, resulting in an imbalanced 3D-versus-2D data ratio. 

\subsection{Animal Family Supervised Contrastive Learning}
\label{sec:sec:const}
Different from human whose SMPL shape parameters come from the same multivariate normal distribution, animals demonstrate at least two levels of shape differences: inter-family level and intra-family level. For example, dogs share similar shapes between each other, yet have distinct shapes from cows. 
To capture such two levels of shape distributions in AniMer, we propose an animal family loss based on supervised contrastive learning~\cite{khosla2020supervised}.  

Specifically, we employ a learnable class token~\cite{dosovitskiy2020image} to represent the animal family. First, this token \(\mathbf{x}_{\text{class}}\) together with a minibatch of images \(I \in \mathbb{R}^{B \times 3 \times H \times W}\) are fed into the ViT encoder, where \(B\) is the batch size. Then, an MLP head is utilized to generate the animal family feature \(\boldsymbol{z}\) from the class token. Finally, we directly apply supervised contrastive learning to \(\boldsymbol{z}\) through a loss $\mathcal{L}_{\text{con}}$ defined as follows:
\begin{equation}
    \label{animal family loss}
    \mathcal{L}_{\text{con}} =   \sum_{i \in I} 
                            \frac{-1}{|P(i)|} 
                            \sum_{p \in P(i)} \frac{\exp{(\boldsymbol{z_i} \cdot \boldsymbol{z_p} / \tau) }}
                            {\sum_{o \in O(i)} \exp{(\boldsymbol{z_i} \cdot \boldsymbol{z_o} / \tau)}}. 
\end{equation}
Within a minibatch, \(P(i)\) represents samples that share the same family label $i$, $O(i)$ represents samples other than $i$. The notation \(|P(i)|\) denotes the size of this set. \(\tau \in \mathbb{R}^{+}\) is a scalar temperature parameter. 

\subsection{The Architecture of AniMer+}
While AniMer provides a robust framework for diverse mammalian families, its SMAL-based foundation cannot accommodate animals with fundamentally different anatomies, such as birds. A naive approach of training a single, shared network on both mammals and birds risks feature interference, where the model might ``average out" the anatomical distinctions we aim to capture. 

To address this challenge and create a truly unified model, we extend AniMer to \textbf{AniMer+}, which features a specialized architecture designed to concurrently handle multiple distinct taxa. The full architecture of AniMer+ is shown in Fig.~\ref{fig:animerplus_pipeline}. We utilize the idea of MoE~\cite{shazeer2017outrageously, jiang2024mixtral, xu2023vitpose++}. Specifically, we partition the second FC layer into two parallel paths: two taxa-specific layers and one taxa-shared layer. The taxa-specific layers act as experts, learning to extract features relevant only to mammals or birds, respectively, while the shared layer processes features from all taxa, which incentivizes it during training to learn generalizable, cross-species representations.

Specifically, in a ViT-MoE block, the output of the first FC layer and the GELU activation function, denoted as \(\mathbf{F}_{\text{fc1}} \in \mathbb{R}^{193 \times 1280}\) (the number of tokens is 193, and the feature dimension is 1280), is processed through the taxa-shared layer to produce \(\mathbf{F}_{\text{taxa-shared}}\). Subsequently, the taxa-specific layer processes the data according to the species of interest to yield \(\mathbf{F}_{\text{taxa-specific}}\). Finally, the block output \(\mathbf{F}_\text{block}\) is obtained by concatenating \(\mathbf{F}_{\text{taxa-shared}}\) and \(\mathbf{F}_{\text{taxa-specific}}\): 
\begin{equation}
\begin{aligned}
    \label{Moe}
     &\mathbf{F}_{\text{taxa-shared}} = FC_{\text{taxa-shared}}(\mathbf{F}_{\text{fc1}}), \\
     &\mathbf{F}_{\text{taxa-specific}} = FC_{\text{taxa-specific}}(\mathbf{F}_{\text{fc1}}), \\
     &\mathbf{F}_{\text{block}}  = \text{concat}(\mathbf{F}_{\text{taxa-shared}}, \mathbf{F}_{\text{taxa-specific}}),
\end{aligned}
\end{equation}
where \(\mathbf{F}_\text{taxa-shared} \in \mathbb{R}^{193 \times 960}\), \(\mathbf{F}_\text{taxa-specific} \in \mathbb{R}^{193 \times 320}\) and \(\mathbf{F}_\text{block} \in \mathbb{R}^{193 \times 1280}\). 
After the final ViT-MoE block, the feature tokens are passed to the decoder, which then branches to predict the distinct parameters for the SMAL (mammalia) and AVES (aves) models, respectively. \revise{Different from the typical MoE where a gating network is necessary to predict which expert to route each token to, we use a simple hard-coded switch based on the input's taxa (mammals or birds), which is more direct and efficient than a learned gate for this problem. }

In this way, our method can effectively accommodate both mammals and birds. However, the existing bird datasets (e.g., CUB~\cite{wah2011caltech}) only provide 2D annotations, and the absence of 3D priors in single-view reconstruction tasks leads to depth ambiguities. In such context, we utilize our dataset generation pipeline (Sec.~\ref{sec:ctrlani3d}) to obtain the first bird dataset with 3D annotations named CtrlAVES3D. By combining AniMer+ with the dataset generation pipeline, our method can effectively accommodate both mammals and birds, with the potential to generalize to any animal group representable by a parametric model.

\subsection{Loss Functions}
To ensure image-aligned reconstruction results, we train AniMer+ using a comprehensive loss function that incorporates various 2D and 3D annotations. We define the main loss function \(\mathcal{L}_{\text{total}}\) as a weighted sum of several loss components, each focusing on different aspects of the model's performance. The main loss function is given by
\begin{equation}
\begin{aligned}
    \label{total_loss}
    \mathcal{L}_{\text{total}} =& \lambda_{\text{3D}} \mathcal{L}_{\text{3D}} + \lambda_{\text{2D}} \mathcal{L}_{\text{2D}} + \lambda_{\text{smal\_prior}} \mathcal{L}_{\text{smal\_prior}} \\
    +& 
    \lambda_{\text{con}} \mathcal{L}_{\text{con}} + \lambda_{\text{aves\_prior}}\mathcal{L}_{\text{aves\_prior}} .
\end{aligned}
\end{equation}
Here, \(\lambda_{\text{3D}}=0.05, \lambda_{\text{2D}}=0.01, \lambda_{\text{smal\_prior}}=0.001, \lambda_{\text{con}}=0.0005\) and \(\lambda_{aves\_prior}=0.002\) are the loss weights. For 3D training data, all losses are used. For samples without 3D annotations, the 3D loss \(\mathcal{L}_{\text{3D}}\) is disabled. For AniMer training utilizing mammalian or avian datasets, we exclude \(\mathcal{L}_{\text{aves\_prior}}\) or \(\mathcal{L}_{\text{smal\_prior}}\), respectively.

\noindent\textbf{3D Loss.} For images annotated with SMAL model parameters \(\beta\) and \(\theta\), we supervise these parameters directly to enable fast convergence. We also supervise the estimated 3D keypoints \(\hat{K}_{3D}\) with ground truth \(K_{3D}\) to achieve better 3D joint localizations. The full 3D loss function is
\begin{equation}
\begin{aligned}
    \label{3D_loss}
    \mathcal{L}_{\text{3D}} =& \lambda_{\beta} ||\hat{\beta} - \beta||_{2}^{2} + \lambda_{\theta} ||\hat{\theta} - \theta||_{2}^{2} \\
    +&
    ||\hat{K}_{3D} - K_{3D}||_{1} + \lambda_{\alpha}||\hat{\alpha}-\alpha||_{1},
\end{aligned}
\end{equation}
where \(\lambda_{\beta}=0.01\) and \(\lambda_{\theta}=0.2\) are loss weights. When the input image is a bird, \(\lambda_{\alpha}=0.04\), otherwise \(\lambda_{\alpha}=0\). \( ||\cdot||_{2}^{2} \) denotes squared L2 norm, while \( ||\cdot||_{1} \) represents L1 norm. 

\noindent\textbf{2D Loss.} Most training data only contain 2D-level annotations such as 2D keypoints or masks. For these data, we supervise 2D keypoints and masks through
\begin{equation}
    \label{2D_loss}
    \mathcal{L}_{\text{2D}} = ||\pi(\hat{K}_{3D}) - K_{2D}||_{1} + \lambda_{M}||\hat{\mathcal{M}}-\mathcal{M}||_2^2,
\end{equation}
where \(\lambda_{M}=2\), \(\hat{\mathcal{M}}\) and \(\mathcal{M}\) are the rendered mask from the predicted mesh and the ground truth mask, respectively. 

\begin{figure*}[ht]
    \centering
    \includegraphics[width=0.95\linewidth]
    {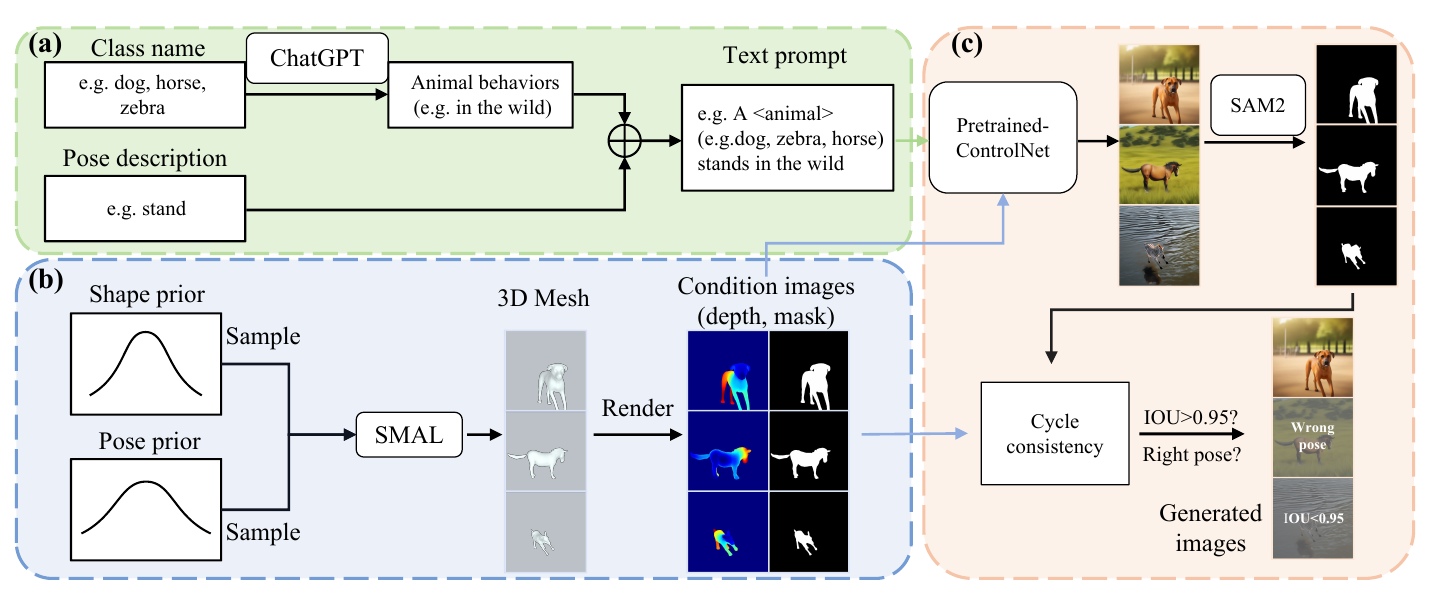}
    \caption{\textbf{Dataset generation pipeline.} The whole pipeline contains three parts: \textbf{(a)} Text prompt generation. 
    \textbf{(b)} Condition image generation. 
    \textbf{(c)} Image generation and post-processing. 
    }
    \label{fig:ctrlani3d_pipeline}
\end{figure*}

\noindent\textbf{SMAL Prior Loss.} For images with only 2D annotations, to ensure that the predicted shape and pose parameters are reasonable, we enforce them to be close to a prior distribution by calculating the Mahalanobis distance. The prior loss is
\begin{equation}
    \label{prior_loss}
    \mathcal{L}_{\text{smal\_prior}} = \lambda_{\beta} (\hat{\beta} - \mu_{\beta})^{T} \Sigma^{-1}_{\beta} (\hat{\beta} - \mu_{\beta}) + (\hat{\theta} - \mu_{\theta})^{T} \Sigma^{-1}_{\theta} (\hat{\theta} - \mu_{\theta}),
\end{equation}
where \(\lambda_{\beta}=0.5\), \(\mu_{\beta}\), \(\Sigma_{\beta}\), \(\mu_{\theta}\), and \(\Sigma_{\theta}\) are the mean and covariance of the prior distributions given by SMAL~\cite{zuffi20173d}.

\noindent\textbf{AVES Prior Loss.} Following the AVES model~\cite{wang2021birds}, the prior loss for aves is as follows:
\begin{equation}
    \label{aves_prior_loss}
    \mathcal{L}_{\text{aves\_prior}}=\lambda_{\beta}||\beta||_2^2+\lambda_{\theta}||\hat{\theta}-\overline{\theta}||_2^2+||\hat{\alpha}-\overline{\alpha}||_2^2,
\end{equation}
where \(\lambda_{\beta}=0.5\) and \(\lambda_{\theta}=1\), \(\overline{\theta}\) and \(\overline{\alpha}\) are the mean pose and mean bone length, respectively.


\section{Dataset Generation}
\label{sec:ctrlani3d}
\noindent\textbf{SMAL Structure Condition for CtrlAni3D.} The generation pipeline for the CtrlAni3D dataset is illustrated in Fig.~\ref{fig:ctrlani3d_pipeline}. Given a posed SMAL mesh and viewpoint, we render it into a mask map and a depth map as the condition images to guide the structure of the images generated by a pre-trained ControlNet~\cite{zhang2023adding}. To guarantee the diversity of the poses and shapes, we randomly sample \(\beta\) from the Gaussian distributed shape space provided by SMAL~\cite{zuffi20173d}, and sample a more diverse range of \(\theta\) from a combined pose space presented by dog~\cite{ruegg2023bite,biggs2020wldo} and horse motion~\cite{li2024poses}. This is reasonable because the quadrupeds expressed by SMAL share similar anatomical structures. About the viewpoint for rendering, each dimension of the global rotation vector is uniformly sampled from $(-\pi,\pi)$ while the position is uniformly sampled between $[-0.5,-0.5,4]$ and $[0.5,0.5,8]$, therefore truncated images may be generated to enhance training's robustness. 

\noindent\textbf{AVES Structure Condition for CtrlAVES3D.} The primary differences in generating CtrlAVES3D compared to CtrlAni3D are \revise{four} folds. 1) Instead of the early version ControlNet utilized by CtrlAni3D, we employ Flux-Union-Pro-V2~\cite{flux-cn-union-pro-2}, a modern version ControlNet that generates more realistic bird textures. 
\revise{2) Due to the lack of a probabilistic distribution of pose or shape space, we randomly sample $\beta$ and $\theta$ from the well-estimated results provided by the original AVES paper~\cite{wang2021birds} on the 297 images of the CUB17 dataset (see Sec.~\ref{sec:sec:setup}).}
3) Except for the shape parameter \(\beta\) and the pose parameter \(\theta\), the AVES model needs an extra bone parameter \(\alpha\) to control the bone lengths. We sample \(\alpha\) within [-0.5, 3.5] according to AVES's practice~\cite{wang2021birds}.
4) We utilize Canny edge and depth map as structural conditions, as Flux-Union-Pro-V2 does not support mask as a condition. 

\noindent\textbf{Text Condition.} To further control the style of the generated images $x$, we seek to use text prompts. During the generation of the animal datasets of interest, we prompt ControlNet using common names instead of scientific names. 
For the SMAL parametric model, we manually classify the sampled 3D meshes into 10 species: cat, tiger, lion, cheetah, dog, wolf, horse, zebra, cow, and hippo, and use the species name as one keyword. For the AVES parametric model, the species name can be found in Table~\ref{tab:taxonomy_bird}. When it is hard to distinguish (e.g., some complicated cases of zebra/horse), we use both animal names as prompts and select the one with the highest cycle consistency. The second keyword is the pose description, e.g.,``stands" in Fig.~\ref{fig:ctrlani3d_pipeline}, which is manually assigned according to the 3D mesh pose. Based on these keywords, ChatGPT~\cite{achiam2023gpt} is employed to automatically complete a prompt sentence $c_t$ depicting possible animal behavior. Finally, both $c_v$ and $c_t$ act as the prompts of ControlNet for realistic and rich image generation.

\begin{table}[ht]
\centering
\caption{{Scientific names (second column) of the quadruped species used in CtrlAni3D. } The image count of each species is tabulated at the rightmost column. }
\resizebox{0.49\textwidth}{!}
{
\begin{tabular}{llll} \toprule
Family                   & Species                & \makecell[c]{Prompt\\ Names} & Count \\ \midrule
\multirow{4}{*}{Felidae} & \textit{Felis catus} & Cat & 80 \\
                  & \textit{Panthera leo} & Lion & 630 \\
                  & \textit{Acinonyx jubatus} & Cheetah & 299 \\
                  & \textit{Panthera tigris  } & Tiger & 280 \\ \hline
\multirow{2}{*}{Canidae} & \textit{Canis lupus familiaris} & Dog & 2976 \\
                  & \textit{Canis lupus } & Wolf & 413 \\ \hline
\multirow{2}{*}{Equidae} & \textit{Equus ferus caballus} & Horse & 2228 \\
                  & \textit{Equus zebra} & Zebra & 1460 \\ \hline
Bovidae  & \textit{Bos taurus}  & Cow & 890 \\ \hline
Hippopotamidae & \textit{Hippopotamus amphibius} & Hippo & 455 \\ \hline
\multicolumn{3}{l}{\textbf{Total}} & \textbf{9711}   \\ \bottomrule
\end{tabular}
}
\label{tab:taxonomy}
\end{table}

\begin{table}[ht]
\caption{{The prompt name and image count of each species in CtrlAVES3D.}}
\resizebox{0.48\textwidth}{!}
{
\begin{tabular}{llll}
\toprule
Prompt Names     & Count & Prompt Names           & Count \\ \hline
Laysan albatross    & 501   & Painted bunting           & 494   \\ \hline
Cardinal            & 447   & American crow             & 414   \\ \hline
Northern flicker    & 441   & Scissor tailed flycatcher & 264   \\ \hline
Boat tailed grackle & 334   & Evening grosbeak          & 363   \\ \hline
California gull     & 394   & Blue jay                  & 370   \\ \hline
Green kingfisher    & 487   & White breasted kingfisher & 501   \\ \hline
Mallard             & 287   & Horned puffin             & 330   \\ \hline
Geococcyx           & 305   & Cedar waxwing             & 467   \\ \hline
Pileated woodpecker & 374   & \textbf{Total}                     & \textbf{6773}  \\ \bottomrule
\end{tabular}
}
\label{tab:taxonomy_bird}
\end{table}

\begin{figure}[ht]
    \centering
    \includegraphics[width=\linewidth]{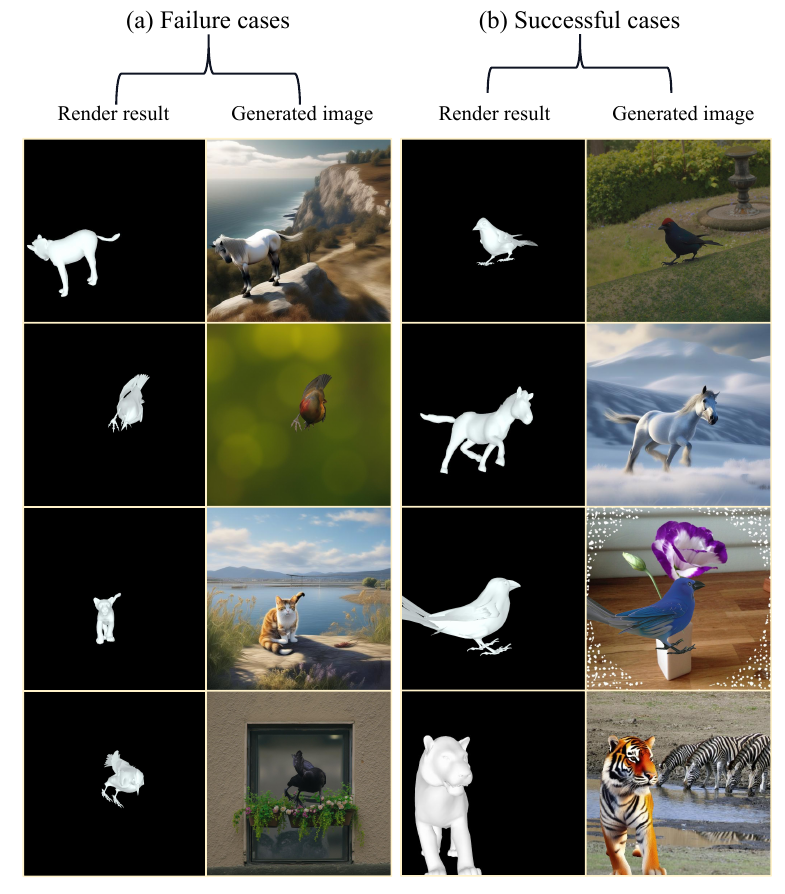}
    \caption{\textbf{CtrlAni3D and CtrlAVES3D failure cases and successful cases.} (a) Failure cases. There are two main cases of failure: (1) At times, ControlNet may struggle to generate mesh-aligned poses (first row and second row). (2) Additionally, ControlNet may not effectively generate the intricate details of the animal body (third row and fourth row). (b) Successful cases. The backgrounds of the first and second rows are generated by ControlNet, while the backgrounds of the third and fourth rows are sourced from the COCO dataset.}
    \label{fig:data_generate_failure_cases}
\end{figure}

\noindent\textbf{Semi-Automated Filtering.} Note that not all generated images align perfectly with the conditions. To address this issue, we develop a semi-automated filtering strategy to reduce the burden on annotators. First, we utilize SAM2~\cite{ravi2024sam2} to extract the foreground mask from each generated image, enabling cycle-consistency checking by comparing it to the mask. Further, we manually filter out images that do not match the mesh poses to ensure high data quality. These misaligned results are denoted as ``Failure cases". Some failure and successful cases are illustrated in Fig.~\ref{fig:data_generate_failure_cases}. Each synthetic image includes well-aligned annotations for the parametric model parameters (\(\beta\), \(\theta\), and \(\gamma\) for SMAL; \(\beta\), \(\theta\), \(\alpha\), and \(\gamma\) for AVES) and 3D keypoints (26 keypoints for SMAL; 18 keypoints for AVES). By comparing the rendered depth image with the projected keypoint depths, we also obtain visible 2D keypoints as annotations. \revise{These are feasible because all the keypoints are defined on the body surface.} Specifically, the visibility of the 2D keypoints is determined by comparing the depth \(d_k\) of each keypoint with the depth \(d_p\) at the corresponding pixel location. Specifically, visibility is assigned a value of 1 when \(d_k \le d_p\); otherwise, it is assigned a value of 0.

\noindent\textbf{Backgrounds.}
To ensure background diversity, we randomly sample COCO~\cite{lin2014microsoft} images to serve as backgrounds for \(2/7\) of the CtrlAni3D images and \(1/4\) of the CtrlAVES3D images. The COCO backgrounds are used only when SAM2 achieves satisfactory segmentation quality. Therefore, increasing the ratio of the COCO backgrounds is equivalent to lowering the IoU threshold, which results in a decline in data quality. This may negatively impact model training. 

%% file: src/04_experiment.tex
\section{Experiments}

\subsection{Setup}
\label{sec:sec:setup}
\noindent\textbf{Mammalian Datasets Summary.}
We curate and aggregate multiple mammalian datasets containing 2D and 3D annotations The full training dataset includes the training split of Animal Pose~\cite{cao2019cross}, APT-36K~\cite{yang2022apt}, AwA-Pose~\cite{banik2021novel}, Stanford Extra~\cite{biggs2020wldo}, Zebra synthetic~\cite{zuffi2019three}, Animal3D~\cite{xu2023animal3d}, and our own CtrlAni3D. Only Animal3D, CtrlAni3D and Zebra synthetic have 3D annotations. For all datasets, we exclude images of animals not included in SMAL~\cite{zuffi20173d}, such as elephants. Furthermore, we map the keypoints' order across different datasets to be the same as that of the Animal3D dataset because the order of the annotated keypoints varies from dataset to dataset. For keypoints that are labeled in the Animal3D dataset but not labeled in a specific dataset, we set the weight of the keypoint to zero. We then aggregate all the aforementioned datasets (excluding Animal Kingdom~\cite{Ng_2022_CVPR}) for training, assigning different sampling weights to each dataset based on its type and size, as shown in Table~\ref{tab:train_data_weights}. For evaluation, we mainly use the test part of Animal3D, CtrlAni3D, and Animal Kingdom. Note that, Animal Kingdom serves as the OOD benchmark because it contains very challenging in-the-wild images which are never seen during training. A total of 8 mammalian species (horse, wolf, cheetah, leopard, dog, lion, jaguarundi cat and snow leopard) from Animal Kingdom are selected for testing.

\noindent\textbf{Avian Datasets Summary.}
We use the the training subset of the CUB dataset~\cite{wah2011caltech} and the training part of our CtrlAVES3D dataset as the training dataset. Note that only CtrlAVES3D has 3D annotations. Other than the CUB and CtrlAVES3D datasets, we also evaluate our model on the Cow Bird dataset~\cite{badger20203d} to demonstrate the generalizability of our model. 

\noindent\textbf{Animal Pose Dataset.} The Animal Pose dataset \cite{cao2019cross} includes 5 categories: dog, cat, cow, horse and sheep, comprising a total of over 6,000 instances across more than 4,000 images. Each animal instance is annotated with 20 keypoints.

\noindent\textbf{APT-36k Dataset.} The APT-36k dataset \cite{yang2022apt} contains 36000 images covering 30 animal species from different scenes. Each animal instance is labeled with 17 keypoints.

\noindent\textbf{AwA Pose Dataset.} The AwA Pose dataset \cite{banik2021novel} is introduced for 2D quadruped animal pose estimation. It contains 10064 images of 35 quadruped animal species and each image is annotated with 39 keypoints.

\noindent\textbf{Stanford Extra Dataset.} The Stanford Extra dataset \cite{biggs2020wldo} consists of 20,580 images and covers 120 dog breeds. Each image is annotated with 20 2D keypoints and silhouette.

\noindent\textbf{Zebra synthetic Dataset.} The Zebra synthetic dataset \cite{zuffi2019three} consists of 12850 images. Each image is randomly generated that differs in background, shape, pose, camera, and appearance.

\noindent\textbf{Animal Kingdom Dataset.} The Animal Kingdom dataset \cite{Ng_2022_CVPR} includes a diverse range of animal species. We only use 8 major mammalian species within the pose estimation dataset to evaluate our method.  

\noindent\textbf{Animal3D Dataset.} The Animal3D dataset \cite{xu2023animal3d} contains a total of 3379 images, which are classified into 40 classes. Each image is annotated with SMAL \cite{zuffi20173d} parameters, 2D keypoints, 3D keypoints and masks.

\noindent\textbf{CUB Dataset.} The CUB dataset~\cite{wah2011caltech} contains 200 subcategories of birds. Each image is labeled with a subcategory label, 15 part locations, 312 binary attributes and a bounding box.

\noindent\textbf{CUB17 Dataset.} The CUB17 dataset~\cite{wang2021birds} contains 17 categories (as listed in Table~\ref{tab:taxonomy_bird}) selected from the CUB dataset and 297 images in total. Different from the CUB dataset, the images in the CUB17 dataset are annotated with 18 keypoints. This dataset is only used for training, and the AVES models estimated by Wang \textit{et al.}~\cite{wang2021birds} on these images are used as the pose and shape space of CtrlAVES3D. 

\noindent\textbf{Cow Bird Dataset.} The Cow Bird dataset~\cite{badger20203d} comprises 1,000 images of 15 brown-headed cowbirds recorded at 40 Hz during a three-month mating season, annotated with 6,355 pixel-wise segmentation masks and corresponding bounding boxes. A subset of 18 moments includes 1,031 mask annotations, each enriched with 12 semantic keypoints per bird.

\noindent\textbf{CtrlAni3D and CtrlAVES3D Datasets.} Our datasets are annotated in the same style as the Animal3D dataset. More details about our dataset can be found in Sec.~\ref{sec:ctrlani3d}.

\begin{table}[htbp]
\centering
\caption{\textbf{Full dataset statistics for training.}}
\resizebox{0.95\columnwidth}{!}
{
\begin{tabular}{lllc}
\toprule
Dataset         & Number & Ratio  & Training Sample Weight \\ \hline
Animal3D        & 3065   & 5.7\%  & 1                      \\
CtrlAni3D      & 8277   & 15.3\% & 0.6                    \\
Animal Pose     & 1680   & 3.1\%  & 0.15                   \\
AwA-Pose        & 2884   & 5.3\%  & 0.15                   \\
Zebra Synthetic & 12850  & 23.8\% & 0.05                   \\
Stanford Extra  & 7689   & 14.2\% & 0.15                   \\
APT-36K         & 4887   &  9.0\% & 0.15                   \\
CtrlAVES3D      & 6464   & 12.0\% & 0.45                     \\
CUB             & 5964   & 11.0\% & 0.45                    \\
\revise{CUB17}           & \revise{297}    & 0.5\%  & 0.45                    \\
Total           & \revise{54057}  & 100\%  & -                      \\ \bottomrule
\end{tabular}
}
\label{tab:train_data_weights}
\end{table}

\noindent\textbf{Metrics.} Multiple 3D and 2D metrics are employed to fully assess the model performance. \textit{PA-MPJPE} is Procrustes-Aligned Mean Per Joint Position Error for 3D keypoints. 
\textit{PA-MPVPE} is Procrustes-Aligned Mean Per Vertex Position Error for SMAL/AVES vertices. 
\textit{PCK} is Percentage of Correct Keypoints within a certain threshold to the ground truth keypoints. Note that the distances between the predicted and ground truth keypoints are normalized by the square root of the 2D silhouette area rather than that of the bounding box area. In this paper, PCK is only used for evaluating 2D keypoints. PCK@HTH uses half the head-to-tail distance as the threshold. By setting the threshold to 0.1 and 0.15, we get commonly used PCK@0.1 and PCK@0.15 metrics. 
\textit{AUC} is Area Under the Curve value when the PCK threshold gradually increases from 0 to 1. Note that, MPJPE without Procrustes-Alignment is not used because large animal size variations would cause imbalanced MPJPE among species. 
\textit{PA-CD} is Chamfer Distance after Procrustes alignment between the 3D ground truth geometry and predicted geometry, which is used for comparing meshes of different topologies. 

\noindent\textbf{Training Details.}
Our model is implemented by PyTorch~\cite{paszke2019pytorch}. We use AdamW~\cite{adamw} optimizer with a linear learning rate decay schedule. The initial learning rate is $1.25 \times 10^{-6}$. For mammalian experiments, we train AniMer-a (replacing the ViT with ResNet152), AniMer-b (no pretraining), AniMer-c (replacing the Transformer decoder with MLP), and AniMer-M (training AniMer on mammalian datasets) for total 2M steps. 
\revise{Specifically, at the first stage, these models are trained for 200K steps (500 epochs) using only 3D datasets to establish a strong 3D prior. At the second stage, these models are then fine-tuned for an additional 1.8M steps (700 epochs) on the aggregated collection of all 3D and 2D datasets. }

For avian experiments, we train AniMer-A (training AniMer on avian datasets) for 240K steps \revise{in a single stage.} 

To directly compare AniMer and AniMer+, we train AniMer-AM (training AniMer on both mammalian and avian datasets) and AniMer+ for 1.4M steps on the same datasets. 
\revise{Except for specifically stated experiments, all other experimental setups are conducted under the aforementioned conditions.}
All experiments above are conducted on a single NVIDIA RTX 4090 GPU.

\begin{table*}[ht]
\caption{\textbf{Quantitative comparisons on the Animal3D, CtrlAni3D, and Animal Kingdom datasets.} \textbf{Bold} and \underline{underlined} numbers indicate the best performance and the second best performance, respectively. P@H, P@0.1, P@0.15, PAJ, and PAV represent PCK@HTH, PCK@0.1, PCK@0.15, PA-MPJPE, and PA-MPVPE, respectively. 
}
\label{table: quadrupeds_comparison_results_to_baseline}
\centering
\resizebox{0.9\textwidth}{!}{
\begin{tabular}{ccccccccccccc}
\toprule
Dataset & \multicolumn{4}{c}{Animal3D} & \multicolumn{4}{c}{CtrlAni3D} & \multicolumn{4}{c}{Animal Kingdom} \\ 
\cmidrule(lr){2-5} \cmidrule(lr){6-9} \cmidrule(lr){10-13} 

Metric  & AUC\(\uparrow\) & P@H\(\uparrow\) & PAJ\(\downarrow\) & PAV\(\downarrow\) & AUC\(\uparrow\) & P@H\(\uparrow\) & PAJ\(\downarrow\) & PAV\(\downarrow\) & AUC\(\uparrow\) & P@H\(\uparrow\) & P@0.1\(\uparrow\) & P@0.15\(\uparrow\) \\ \hline

HMR    & 76.3 & 60.8 & 123.5 & 133.9 & 80.8 & 67.0 & 123.5 & 133.9 & 70.2 & 64.0 & 12.8 & 25.6 \\
WLDO   & 78.2 & 68.7 & 112.3 & 125.2 & 88.7 & 86.7 & 71.5 & 83.4 & 70.1 & 64.3 & 14.6 & 27.6 \\
AniMer-a & 75.2 & 57.2 & 115.5 & 128.7 & 80.3 & 66.0 & 117.0 & 129.4 & 68.9 & 62.5 & 10.2 & 21.3 \\
AniMer-b & 60.6 & 38.9 & 147.9 & 157.6 & 78.5 & 65.9 & 102.3 & 112.6 & 45.4 & 31.8 & 4.0 & 9.2 \\
HMR2.0 &  86.7 &  84.6 &  94.1 &  98.5 &  91.8 &  93.0 & 60.9 &  66.4 &  77.3 &  73.9 &  22.7 &  40.2 \\
\midrule
AniMer-M  & \textbf{88.9} & \underline{89.5} & \underline{80.4} & \underline{85.7} & \textbf{93.8} & \underline{95.4} & \textbf{44.1} & \textbf{47.6} & \textbf{82.9} & \underline{83.7} & \textbf{34.9} & \textbf{54.7} \\ 
AniMer+ & \textbf{88.9} & \textbf{93.0} & \textbf{78.0} & \textbf{82.7} & \underline{93.6} & \textbf{95.9} & \underline{47.8} & \underline{50.8} & \underline{82.6} & \textbf{92.9} & \underline{34.2} & \underline{53.0} \\
\bottomrule
\end{tabular}
}
\end{table*}

\begin{table}[htbp]
\caption{\textbf{Quantitative comparisons on the Animal3D dataset with 3D Fauna~\cite{li2024learning}. } PA-CD: Chamfer Distance after Procrustes alignment between the 3D grouth truth and predicted results. Results for PCK@0.1 of 3D Fauna are borrowed from the original paper.}
\label{tab: quantitative_comparison_with_fauna}
\centering
\resizebox{0.6\columnwidth}{!}{%
\begin{tabular}{ccc}
\toprule
Method   & PCK@0.1 \(\uparrow\) & PA-CD \(\downarrow\)          \\ \hline
3D Fauna    & 90.1        & 26.39         \\
AniMer-M & \textbf{91.2}  & \textbf{8.39} \\
AniMer+  & 90.0           & 8.60        \\ \bottomrule
\end{tabular}
}
\end{table}

\begin{table*}[ht]
\caption{\textbf{Quantitative comparisons on the CtrlAVES3D, CUB, and Cow Bird datasets.} 
\textbf{Bold} and \underline{underlined} numbers indicate the best performance and the second best performance, respectively. P@H, P@0.1, P@0.15, PAJ, and PAV represent PCK@HTH, PCK@0.1, PCK@0.15, PA-MPJPE, and PA-MPVPE, respectively. 
}
\label{table: birds_comparison_results_to_baseline}
\centering
\resizebox{0.9\textwidth}{!}{
\begin{tabular}{ccccccccccccc}
\toprule
Dataset & \multicolumn{4}{c}{CtrlAVES3D} & \multicolumn{4}{c}{CUB} & \multicolumn{4}{c}{Cow Bird} \\ 
\cmidrule(lr){2-5} \cmidrule(lr){6-9} \cmidrule(lr){10-13} 

Metric  & AUC\(\uparrow\) & P@H\(\uparrow\) & PAJ\(\downarrow\) & PAV\(\downarrow\) & AUC\(\uparrow\) & P@H\(\uparrow\) & P@0.1\(\uparrow\) & P@0.15\(\uparrow\) & AUC\(\uparrow\) & P@H\(\uparrow\) & P@0.1\(\uparrow\) & P@0.15\(\uparrow\) \\ \hline
AVES    & 85.9 & 88.3 & 86.5 & 93.9 & 79.7 & 88.6 & 25.2 & 45.7 & 46.1 & 27.2 & 5.6 & 11.7 \\
HMR2.0 &  91.6 &  95.3 &  68.4 &  90.3 &  87.7 &  96.8 &  50.0 &  73.2 &  59.1 &  45.9 &  12.2 &  24.4 \\
\midrule 
AniMer-A  & \underline{92.9} & \textbf{96.4} & \textbf{65.5} & \underline{90.2} & \textbf{91.9} & \textbf{98.7} & \textbf{74.8} & \textbf{90.2} & \textbf{65.7} & \textbf{56.7} & \textbf{19.4} & \textbf{31.3} \\ 
AniMer+ & \textbf{93.0} & \underline{96.3} & \underline{65.6} & \textbf{70.9} & \underline{89.5} & \underline{97.7} & \underline{59.4} & \underline{80.5} & \underline{59.3} &  \underline{47.2} & \underline{16.4} & \underline{28.4}  \\
\bottomrule
\end{tabular}
}
\end{table*}



\begin{figure*}[htbp]
    \centering
     \includegraphics[width=1.0\linewidth]{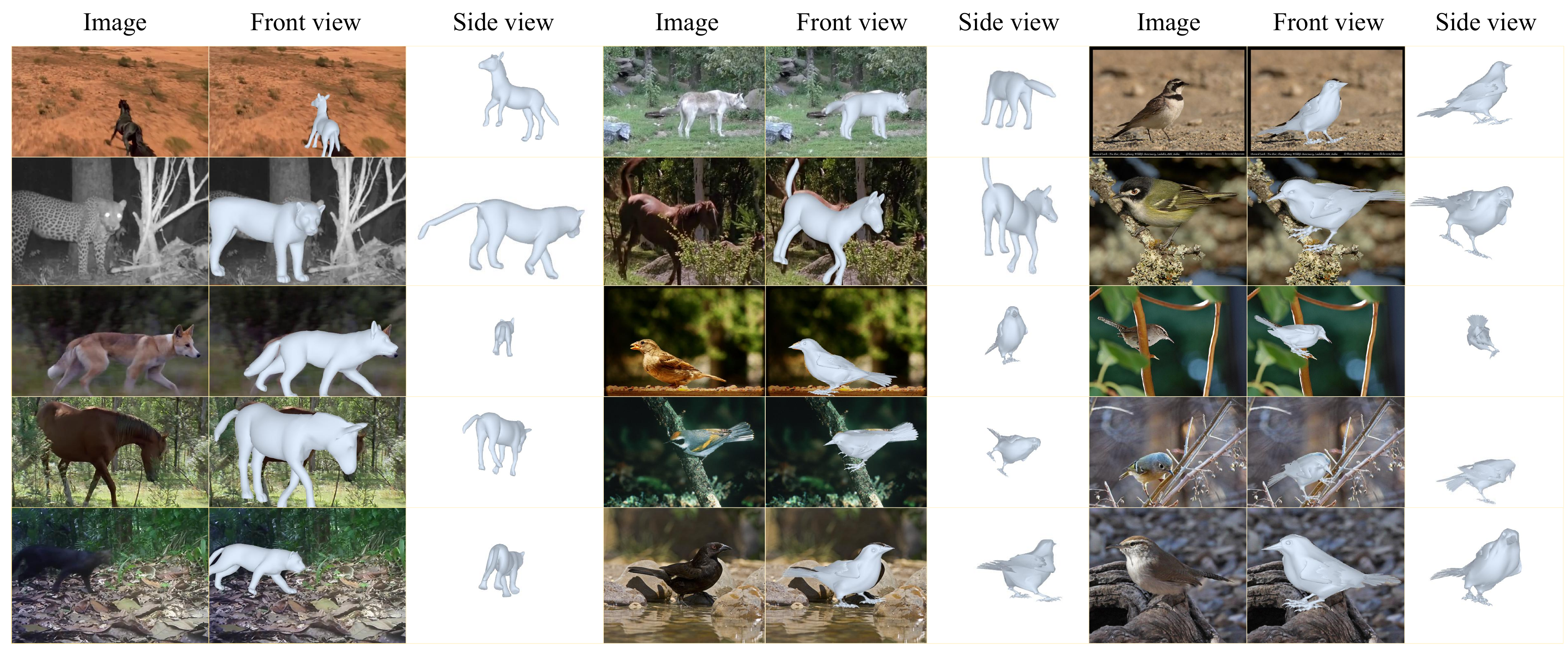}
    \caption{\textbf{Qualitative results of AniMer+ on Animal Kingdom and CUB datasets.} For each case, we display the input image and the output result, which include both a front view rendering and a side view rendering. The mammals and birds are sourced from the Animal Kingdom dataset and the CUB dataset, respectively. 
    }
    \label{fig:qualitative_animalplus}
\end{figure*}

\subsection{Qualitative Results of AniMer+}
We provide qualitative results on the OOD Animal Kingdom dataset and the CUB dataset in Fig.~\ref{fig:qualitative_animalplus}. AniMer+ can estimate pose and shape for both quadrupeds and birds through a single network. As can be seen in Fig.~\ref{fig:qualitative_animalplus}, AniMer+ performs well even in challenging conditions such as motion blur, unusual lighting, partial occlusion and truncation.

\begin{figure*}[ht]
    \centering
     \includegraphics[width=0.95\linewidth]{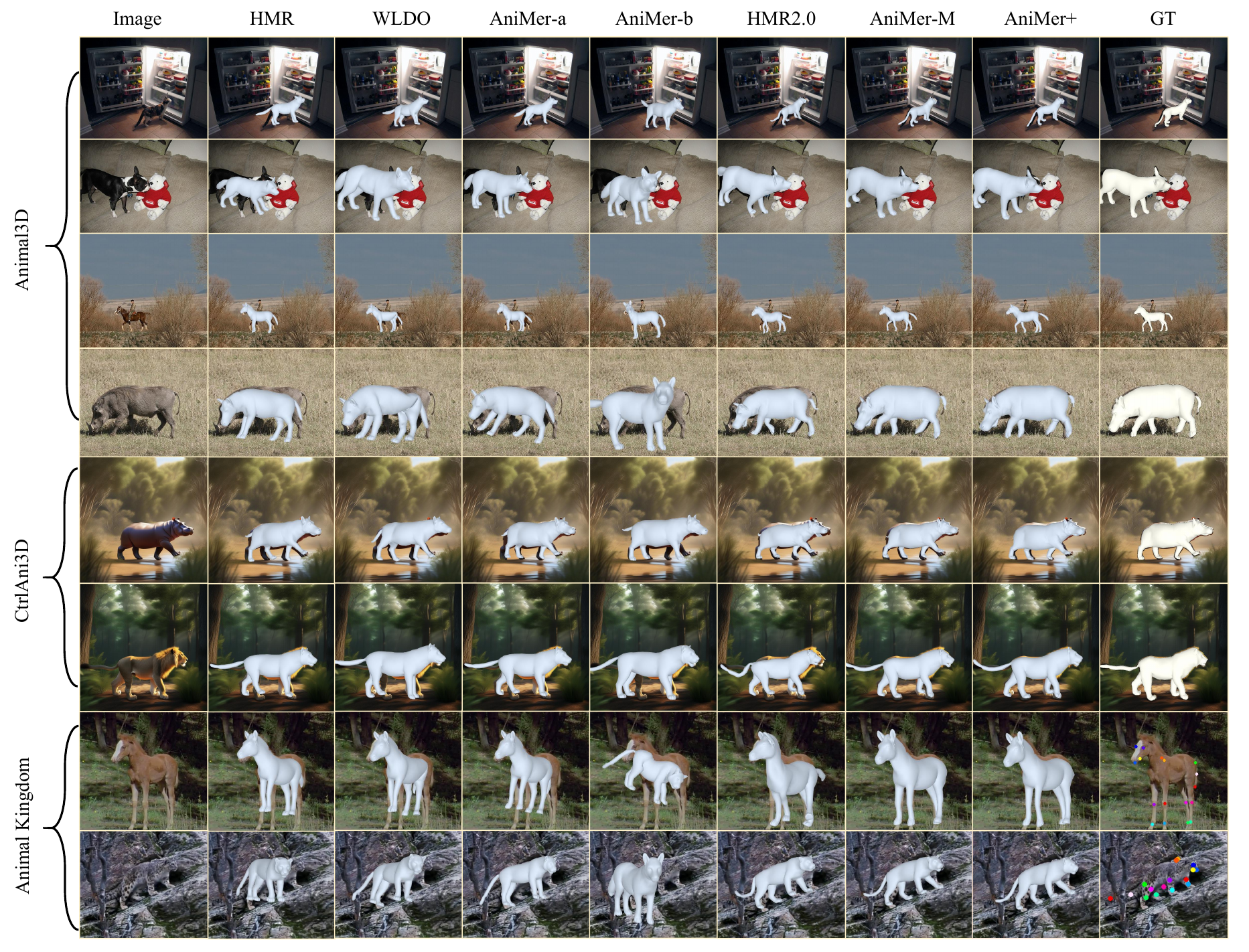}
    \caption{\textbf{Qualitative comparisons on the Animal3D, CtrlAni3D, and Animal Kingdom datasets.} We compare our results with HMR~\cite{kanazawa2018end}, WLDO~\cite{biggs2020wldo}, AniMer-a (ResNet152 backbone), AniMer-b (no pretraining) and HMR2.0~\cite{goel2023humans}. 
    }
    \label{fig:quali_compare_main}
\end{figure*}

\begin{figure}[htbp]
    \centering
     \includegraphics[width=1.0\linewidth]{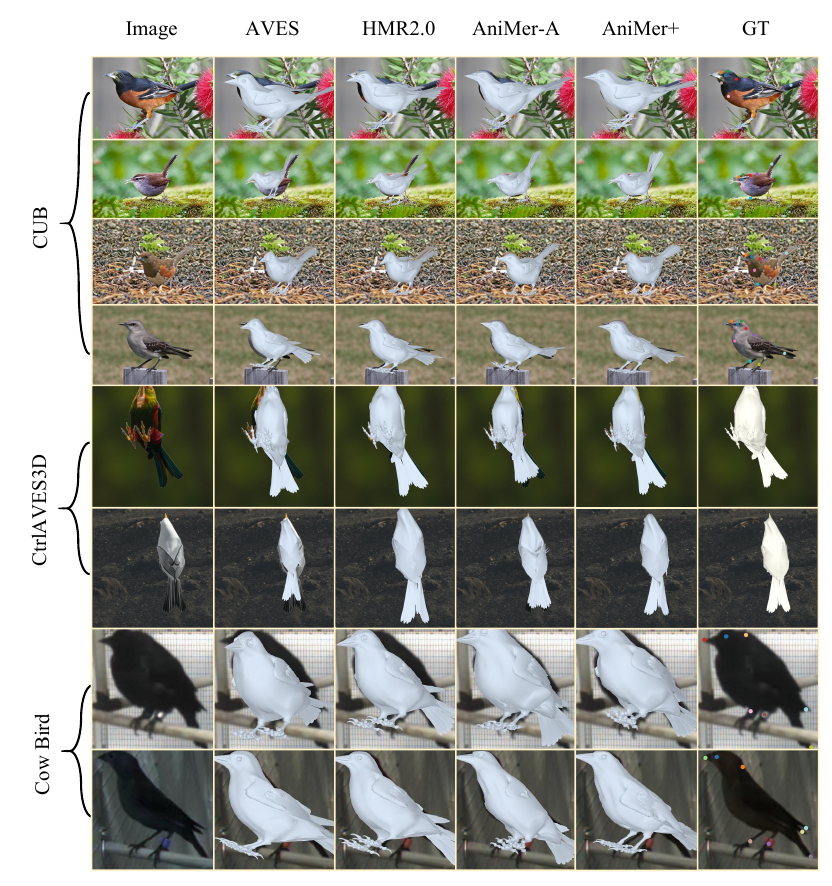}
    \caption{\textbf{Qualitative comparisons on the CUB, CtrlAVES3D and Cow Bird datasets.} We compare our results with AVES~\cite{wang2021birds} and HMR2.0~\cite{goel2023humans}. 
    }
    \label{fig:quali_compare_aves}
\end{figure}

\subsection{Comparisons}
\subsubsection{Mammalian Experiments}
Following the practice of the Animal3D dataset~\cite{xu2023animal3d}, we utilize HMR~\cite{kanazawa2018end} and WLDO~\cite{biggs2020wldo} as baselines. Also, we compare with HMR2.0~\cite{goel2023humans} to highlight our special design choices. To be fair, all compared methods are trained on our full training dataset with the same losses and the same two-stage training strategy.   

Quantitative results are shown in Table~\ref{table: quadrupeds_comparison_results_to_baseline}. We observe that AniMer-M and AniMer+ consistently outperform previous works across multiple datasets in terms of both 3D and 2D metrics.
Specifically on the Animal Kingdom dataset, our method demonstrates strong robustness and improves over previous works in terms of all metrics. 
Qualitative results in Fig.~\ref{fig:quali_compare_main} show that our method aligns with images much better for thin structures such as legs and tails. 

In addition, we compare our method with a typical template-free method 3D Fauna~\cite{li2024learning} on the Animal3D dataset. The PCK@0.1 results for 3D Fauna are sourced from the original paper. Note that, the PCK@0.1 of 3D Fauna uses square root of the 2D bounding box area as a normalization factor. For fair comparison, we use the same threshold for PCK@0.1 calculation in this experiment. Besides, due to the inconsistency between the mesh topology of the 3D Fauna output and SMAL, we utilize PA-CD as the 3D evaluation metric instead of PA-MPJPE and PA-MPVPE. The quantitative results are displayed in Table~\ref{tab: quantitative_comparison_with_fauna}. AniMer-M and AniMer+ demonstrate comparable performance to 3D Fauna on PCK@0.1. For PA-CD, AniMer-M and AniMer+ show improvements of \(68.2\%\) and \(67.4\%\), respectively, compared to 3D Fauna.

\subsubsection{Avian Experiments}
\label{sec:sec:sec:avian_exp}
We conduct both quantitative and qualitative comparisons against AVES~\cite{wang2021birds} and HMR2.0~\cite{goel2023humans}. Unlike mammals, birds exhibit a more limited range of poses and shapes, so the model converges faster. Therefore, we train AniMer-A in a single stage. The quantitative results in Table~\ref{table: birds_comparison_results_to_baseline} show that AniMer-A and AniMer+ outperform all baselines across all datasets. Notably, our method achieves the best results on the Cow Bird dataset despite its large number of blurred images and the fact that this dataset is not seen during training—underscoring the robustness of our method. 
While AniMer+ and AniMer-A perform comparably on our synthetic CtrlAVES3D dataset, the specialized AniMer-A achieves higher accuracy on the real-world CUB and Cow Bird datasets, as its entire network capacity is dedicated to avian reconstruction. 

The qualitative evaluations in Fig.~\ref{fig:quali_compare_aves} illustrate that our method produces more precise alignments with the input images in terms of avian structure, a benefit we attribute to our family‑aware training strategy.

\subsection{Ablation Study on Datasets}
\subsubsection{Ablations on CtrlAni3D}
To further demonstrate the effectiveness of our proposed CtrlAni3D dataset in enhancing network capability, we conduct a series of ablation experiments. 
We compare AniMer-M trained on different combinations of Animal3D (A3D), CtrlAni3D (C3D), and other datasets (mammalian datasets except A3D and C3D). The hyperparameter training settings remain unchanged.
\begin{table}[bp]
\centering
\begin{small}
\caption{\textbf{Effect of CtrlAni3D (C3D) on the OOD Animal Kingdom dataset.} 
We report AUC, PCK@0.1 and PCK@0.15 as 2D metrics. \textbf{Bold} numbers indicate the best performance.}
\label{tab:ablation_2D}
\resizebox{0.9\columnwidth}{!}{%
\begin{tabular}{ccc ccc}
\toprule
\multicolumn{3}{c}{Training Data} & \multicolumn{3}{c}{2D Metric} \\
\cmidrule(lr){1-3} \cmidrule(lr){4-6}
A3D & C3D & others & AUC$\uparrow$ & PCK@0.1$\uparrow$ & PCK@0.15$\uparrow$ \\
\midrule
    &     & \dui   & 78.5  & 24.5 & 44.9  \\
    & \dui & \dui   & 82.7  & 33.8 & 53.2  \\
 \dui & \dui &       & 81.9  & 32.1 & 51.8  \\
 \dui &     & \dui   & 82.1  & 33.7 & 53.2  \\
 \dui & \dui & \dui   & \textbf{82.9} & \textbf{34.9} & \textbf{54.7} \\
\bottomrule
\end{tabular}%
}
\end{small}
\end{table}

\begin{table}[ht]
\caption{\textbf{Effect of including CtrlAni3D in training.} We evaluate the performance of 3D pose estimation on two models. For the first model, we do not use CtrlAni3D during training, while for the second model, we incorporate CtrlAni3D into the training process.}
\centering
\resizebox{0.8\columnwidth}{!}{%
\label{tab: effect of including CtrlAni3D in training}
\begin{tabular}{ccc}
\toprule
                       & PA-MPJPE \(\downarrow\) & PA-MPVPE \(\downarrow\)  \\ \hline
w/o CtrlAni3D   & 82.6     & 88.4     \\ \hline
w/ CtrlAni3D & \textbf{80.4}    & \textbf{85.7}     \\ \bottomrule
\end{tabular}%
}
\end{table}

\begin{table}[bp]
\centering
\begin{small}
\caption{\textbf{Effect of CA3D (CtrlAVES3D) on the OOD Cow Bird dataset.} 
We report AUC, PCK@HTH, PCK@0.1 and PCK@0.15 as 2D metrics. \textbf{Bold} numbers indicate the best performance.}
\label{tab:ablation_2D_ctrlani3dplus}
\resizebox{0.9\columnwidth}{!}{%
\begin{tabular}{c@{\hskip 2pt}c@{\hskip 2pt}c@{\hskip 2pt}c@{\hskip 2pt}c@{\hskip 2pt}c}
\toprule
\multicolumn{2}{c}{Training Data} & \multicolumn{4}{c}{2D Metric} \\
\cmidrule(lr){1-2} \cmidrule(lr){3-6}
CA3D & CUB & AUC$\uparrow$ & PCK@HTH$\uparrow$ & PCK@0.1$\uparrow$ & PCK@0.15$\uparrow$ \\
\midrule
     & \dui & 53.4 & 39.3 & 11.3 & 19.9 \\
 \dui &      & 62.7 & 51.3 & 16.6 & 30.9 \\
 \dui & \dui & \textbf{65.7} & \textbf{56.7} & \textbf{19.4} & \textbf{31.3} \\
\bottomrule
\end{tabular}%
}
\end{small}
\end{table}

\begin{figure}[htbp]
    \centering
    \includegraphics[width=1.0\linewidth]{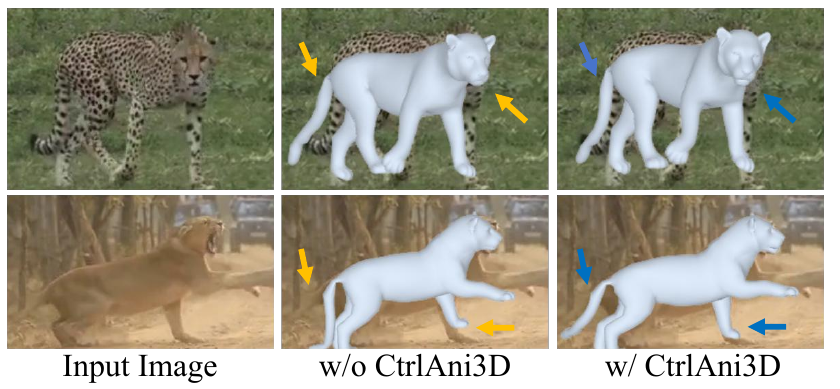}
    \caption{\textbf{Effect of CtrlAni3D on the Animal Kingdom dataset.} Input and result images are zoom-in cropped for visualization. Training with CtrlAni3D enhances the model's ability to align tails, limbs and faces. Orange arrows indicate misalignments, while blue arrows indicate better alignments.}
    \label{fig:without_ctrlani3d}
\end{figure}

Specifically, Table~\ref{tab:ablation_2D} indicates that our model, when trained with CtrlAni3D, exhibits superior performance even on previously unseen in-the-wild data. Qualitative comparisons in 
Fig.~\ref{fig:without_ctrlani3d} further demonstrate that training with CtrlAni3D helps AniMer to yield more accurate animal terminal body parts such as tails, limbs and faces. 

Additionally, to demonstrate the improvements of CtrlAni3D for 3D pose estimation, we report results of the 3D metrics on Animal3D, as shown in Table~\ref{tab: effect of including CtrlAni3D in training}. We observe that training with CtrlAni3D enhances performance on Animal3D, which indicates the pose diversity of our CtrlAni3D dataset.

\subsubsection{Ablations on CtrlAVES3D}
Similar to CtrlAni3D, we conduct an experiment to demonstrate the effectiveness of incorporating CtrlAVES3D training for out-of-domain datasets. The results are shown in Table~\ref{tab:ablation_2D_ctrlani3dplus}. Since the Cow Bird dataset includes a variety of viewpoints, and CtrlAVES3D encompasses diverse perspectives and 3D annotations, models trained with CtrlAVES3D outperform those trained without it. When compared to training solely on CUB or CtrlAVES3D, training on both CUB and CtrlAVES3D results in improvements of \(18.7\%\) and \(4.6\%\) in AUC, respectively.

\subsubsection{\revise{Ablations on Filtering Strategy}}
\revise{
The filtering strategy directly contributes to the quality of our synthetic datasets, which in turn significantly influences the model's performance and generalization.  
We compare training with failure cases (``w/ failure cases'' in Table~\ref{tab:ablation_failure_cases}) and training without failure cases (``w/o failure cases'' in Table~\ref{tab:ablation_failure_cases}) on single animal family {Equidae} (horse and zebra in our case).The Equidae subset of CtrlAni3D contains totally 3138 successful cases for horse and zebra. We further purposely generate 1326 failure cases. Both models are trained for 500 epochs to converge. The substantially improved results in Table~\ref{tab:ablation_failure_cases} demonstrate the effectiveness of the filtering strategy. 

\begin{table*}[ht]
\caption{\revise{\textbf{The impact of training with failure cases.} The results are obtained by evaluating on the Animal3D testset.}}
\label{tab:ablation_failure_cases}
\centering
\resizebox{0.88\textwidth}{!}{
\begin{tabular}{cccccccc}
\toprule
\multicolumn{4}{c}{Horse} & \multicolumn{4}{c}{Zebra} \\ \cmidrule(lr){1-4} \cmidrule(lr){5-8}
\multicolumn{2}{c}{w/o failure cases} & \multicolumn{2}{c}{w/ failure cases} & \multicolumn{2}{c}{w/o failure cases}    & \multicolumn{2}{c}{w/ failure cases} \\ \cmidrule(lr){1-2} \cmidrule(lr){3-4} \cmidrule(lr){5-6} \cmidrule(lr){7-8}
PA-MPJPE\(\downarrow\)       & PA-MPVPE\(\downarrow\)       & PA-MPJPE\(\downarrow\)  & PA-MPVPE\(\downarrow\)  & PA-MPJPE\(\downarrow\)       & PA-MPVPE\(\downarrow\)       & PA-MPJPE\(\downarrow\)  & PA-MPVPE\(\downarrow\)  \\ \hline
\textbf{107.9} & \textbf{118.3} & 141.8     & 141.9     & \textbf{117.0} & \textbf{119.6} & 145.6     & 165.1     \\ \bottomrule
\end{tabular}
}
\end{table*}

}

\subsubsection{\revise{Experiments on Challenging Poses}}
\revise{ 
Synthetic pose and shape datasets may encounter a sim-to-real gap, especially for pose diversity. To test the efficacy of our datasets for rare and challenging poses, we manually select 310 images with challenging poses (e.g., lying down) from the Animal Kingdom dataset to validate Animer-M models trained with or without CtrlAni3D (``w/ CtrlAni3D'' vs. ``w/o CtrlAni3D'' in Table~\ref{tab:ablation_pose_bias_c3d}). Note that, we choose the Animal Kingdom dataset because their labels do not relate to SMAL, and therefore the challenging poses in this dataset are more likely to stay outside the sampled pose space of SMAL. The results in Table~\ref{tab:ablation_pose_bias_c3d} prove that CtrlAni3D helps reconstruct rare poses better, even if sampling is not specifically conducted for challenging postures. 

\begin{table}[ht]
\caption{\revise{\textbf{The impact of CtrlAni3D for challenging poses.}  The results are obtained by evaluating on a subset of 310 challenging images from the Animal Kingdom dataset.}}
\label{tab:ablation_pose_bias_c3d}
\centering
\resizebox{0.42\textwidth}{!}{
\begin{tabular}{cccc}
\toprule
Method        & AUC\(\uparrow\)                                   & PCK@0.1\(\uparrow\)                               & PCK@0.15\(\uparrow\)                              \\ \hline
w/o CtrlAni3D & 80.6          & 29.6          & 48.0          \\
w/ CtrlAni3D  & \textbf{81.0} & \textbf{30.4} & \textbf{50.2} \\ \bottomrule
\end{tabular}
}
\end{table}
}

\subsubsection{\revise{Experiments on Dataset Size of CtrlAni3D}}
\revise{ 
The current training sizes of CtrlAni3D (8,277 images) and CtrlAVES3D (6,464 images) are primarily determined by trading off computational resources, generation time, and observed performance during preliminary experiments. To investigate the behavior of the model's performance as the synthetic data size scales, we progressively increase the size of CtrlAni3D from 2.5K to 10K for training with 500 epochs and test on the Animal3D test set. Please note for this scalability analysis, only CtrlAni3D is utilized as the training data. Table~\ref{tab:ablation_scalability} clearly demonstrates a positive correlation between synthetic dataset size and model performance. However, when the dataset size exceeds 5K, the performance gain slows down.  

\begin{table}[ht]
\caption{\revise{\textbf{Dataset scalability analysis on CtrlAni3D.} The results are obtained by training on CtrlAni3D and testing on the Animal3D testset.}}
\label{tab:ablation_scalability}
\centering
\resizebox{0.24\textwidth}{!}{
\begin{tabular}{ccc}
\toprule
Size                    & AUC\(\uparrow\)           & PA-MPJPE\(\downarrow\)       \\ \hline
2.5K                    & 79.5          & 113.5          \\
5K                      & 82.3          & 107.1          \\
7.5K                    & 82.8          & 106.9          \\
10K                     & \textbf{83.5} & \textbf{106.7} \\ \bottomrule
\end{tabular}
}
\end{table}
}

\subsection{Ablations on Model Design}
\subsubsection{Ablations on Different Encoder and Decoder} To emphasize the significance of the ViT encoder, backbone pretraining and the Transformer decoder, we substitute the ViT encoder with a ResNet-152 encoder, discard ViT pretraining and replace the Transformer decoder with an MLP decoder, respectively. The results are presented in Table~\ref{tab: ablation on different encoder and decoder}. Compared with traditional methods, which use CNN as a backbone and an MLP head as a decoder, our ViT encoder and Transformer decoder improve the performance of the model's output mesh and image alignment. In addition, a backbone with random initialization but no pretraining leads to poor performance and is hard to converge.

\begin{table}[htbp]
\caption{\textbf{Ablations on different encoder and decoder.} AniMer-a: use ResNet-152 as the encoder. AniMer-b: discard ViT pretraining. AniMer-c: use MLP as the decoder. PAJ: PA-MPJPE, PAV: PA-MPVPE, P@0.1: PCK@0.1.}
\resizebox{\columnwidth}{!}{%
\label{tab: ablation on different encoder and decoder}
\begin{tabular}{ccccccc}
\toprule
\multirow{2}{*}{Method} & \multicolumn{2}{c}{Animal3D}  & \multicolumn{2}{c}{CtrlAni3D} & \multicolumn{2}{c}{Animal Kingdom} \\ \cline{2-7} 
         & PAJ \(\downarrow\) & PAV \(\downarrow\) & PAJ \(\downarrow\) & PAV \(\downarrow\) & AUC \(\uparrow\)  & P@0.1 \(\uparrow\) \\ \hline
AniMer-a & 115.5 & 128.7 & 117.0 & 129.4 & 68.9 & 10.2  \\
AniMer-b & 147.9 & 157.6 & 102.3 & 112.6 & 45.4 & 4.0 \\
AniMer-c & 83.9 & 89.2 & 55.8 & 60.9 & 81.9 & 31.6  \\
AniMer-M  & \textbf{80.4} & \textbf{85.7} & \textbf{44.1} & \textbf{47.6} & \textbf{82.9}    & \textbf{34.9}   \\ \bottomrule
\end{tabular}%
}
\end{table}

\subsubsection{Ablations on Animal Family Supervised Contrastive Learning}
\begin{table}[htbp]
\caption{\textbf{Effect of animal family supervised contrastive learning.} 
PAJ: PA-MPJPE. PAV: PA-MPVPE. P@0.1: PCK@0.1.}
\label{tab: Effective of Animal family supervised contrastive learning}
\resizebox{\columnwidth}{!}{%
\begin{tabular}{ccccccc}
\toprule
\multirow{2}{*}{AniMer-M} & \multicolumn{2}{c}{Animal3D} & \multicolumn{2}{c}{CtrlAni3D} & \multicolumn{2}{c}{Animal Kingdom} \\ \cmidrule(lr){2-3} \cmidrule(lr){4-5} \cmidrule(lr){6-7}
      & PAJ\(\downarrow\) & PAV\(\downarrow\) & PAJ\(\downarrow\) & PAV\(\downarrow\) & AUC\(\uparrow\) & P@0.1\(\uparrow\) \\ \hline
w/o \(\mathcal{L}_{\text{con}}\) 
& 81.3 & 86.7 & 44.7 & 48.4 & 82.7 & 34.4    \\ 
w/ \(\mathcal{L}_{\text{con}}\)  
& \textbf{80.4} & \textbf{85.7} & \textbf{44.1} & \textbf{47.6} & \textbf{82.9} & \textbf{34.9}   \\ 
\bottomrule

\toprule
\multirow{2}{*}{AniMer-A} & \multicolumn{2}{c}{CtrlAVES3D} & \multicolumn{2}{c}{CUB} & \multicolumn{2}{c}{Cow Bird} \\ \cmidrule(lr){2-3} \cmidrule(lr){4-5} \cmidrule(lr){6-7}
      & PAJ\(\downarrow\) & PAV\(\downarrow\) & AUC\(\uparrow\) & P@0.1\(\uparrow\) & AUC\(\uparrow\) & P@0.1\(\uparrow\) \\ \hline
w/o \(\mathcal{L}_{\text{con}}\) 
& 72.8 & 94.5 & 91.6 & 73.2 & 61.6 & 15.8  \\ 
w/ \(\mathcal{L}_{\text{con}}\)  
& \textbf{65.5} & \textbf{90.2} & \textbf{91.9} & \textbf{74.8} & \textbf{65.7} & \textbf{19.4}   \\
\bottomrule
\end{tabular}%
}
\end{table}

We conduct experiments to demonstrate the effect of animal family supervised contrastive learning. Table~\ref{tab: Effective of Animal family supervised contrastive learning} presents the quantitative results on both mammalian and avain datasets. Clearly, AniMer-M trained with \(\mathcal{L}_{\text{con}}\) outperforms the model that does not utilize \(\mathcal{L}_{\text{con}}\). The improved PA-MPVPE and PA-MPJPE metrics indicate that \(\mathcal{L}_{\text{con}}\) enhances the model's capability in shape and pose estimation.

\begin{figure}[ht]
    \centering
    \includegraphics[width=0.9\linewidth]{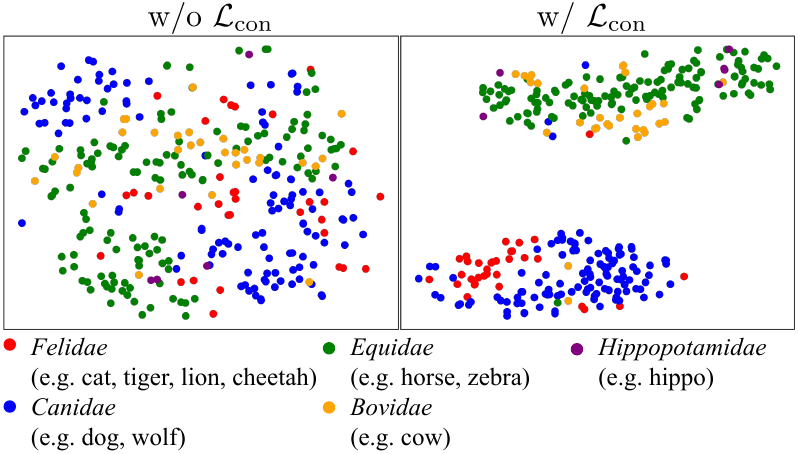}
    \caption{\textbf{t-SNE visualization of the class token on the Animal3D test set.} With the inclusion of \(\mathcal{L}_\text{con}\), features of animals from the same family are more closely clustered.}
    \label{fig:tsne_anisupcon}
\end{figure}

\begin{figure}[htbp]
    \centering
    \includegraphics[width=1.0\linewidth]{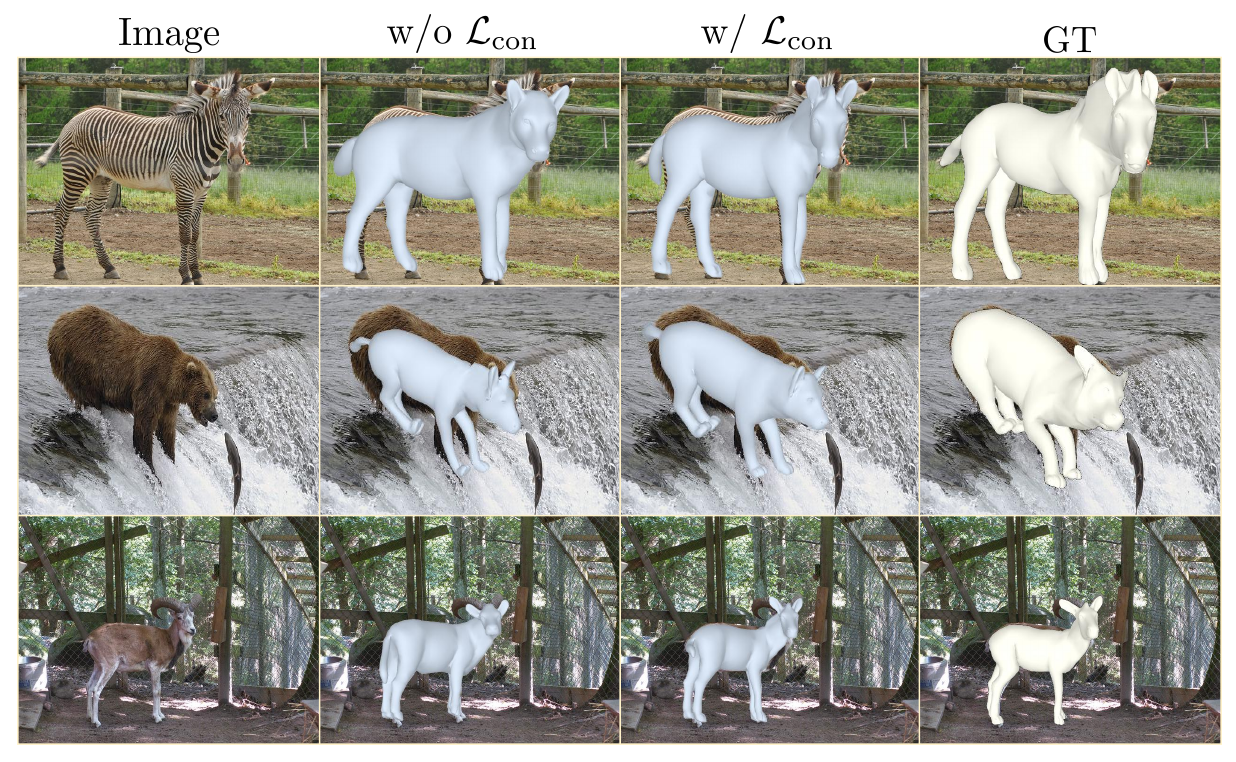} 
    \caption{\textbf{Effect of animal family supervised contrastive learning on the Animal3D dataset.} Without animal family supervised contrastive learning, the estimation is prone to incorrect shapes for animal families.} 
    \label{fig:wo_anisupcon}
\end{figure}

In addition, to provide more intuitive understanding of the effect of 
\(\mathcal{L}_{\text{con}}\), Fig.~\ref{fig:tsne_anisupcon} presents t-SNE~\cite{van2008visualizing} visualization of the class token on the Animal3D test set. It is evident that \(\mathcal{L}_{\text{con}}\) brings features of animals from the same family closer, thereby enhancing the model's ability to distinguish between the shapes of animals from different families. Fig.~\ref{fig:wo_anisupcon} further demonstrates that \(\mathcal{L}_{\text{con}}\) enables the model to better distinguish between different animal families. Specifically, when trained without \(\mathcal{L}_{\text{con}}\), the model incorrectly predicts a zebra, a bear, and an antelope as a dog, a zebra, and a cow, respectively.

\subsubsection{Ablations on AniMer+}
\begin{table}[ht]
\caption{\textbf{Effect of MoE.} PAJ: PA-MPJPE. PAV: PA-MPVPE. P@0.1: PCK@0.1. AniMer-AM: training without MoE. AniMer+: training with MoE.}
\label{tab: Effective of MoE}
\resizebox{\columnwidth}{!}{%
\begin{tabular}{ccccccc}
\toprule
\multirow{2}{*}{Method} & \multicolumn{2}{c}{Animal3D} & \multicolumn{2}{c}{CtrlAni3D} & \multicolumn{2}{c}{Animal Kingdom} \\ \cmidrule(lr){2-3} \cmidrule(lr){4-5} \cmidrule(lr){6-7}
      & PAJ\(\downarrow\) & PAV\(\downarrow\) & PAJ\(\downarrow\) & PAV\(\downarrow\) & AUC\(\uparrow\) & P@0.1\(\uparrow\) \\ \hline
AniMer-AM
& 79.1 & 84.0 & 47.9 & 50.9 & 82.2 & 32.5 \\ 
AniMer+
& \textbf{78.0} & \textbf{82.7} & \textbf{44.1} & \textbf{47.6} & \textbf{82.6} & \textbf{33.1}   \\ 
\bottomrule

\toprule
\multirow{2}{*}{Method} & \multicolumn{2}{c}{CtrlAVES3D} & \multicolumn{2}{c}{CUB} & \multicolumn{2}{c}{Cow Bird} \\ \cmidrule(lr){2-3} \cmidrule(lr){4-5} \cmidrule(lr){6-7}
      & PAJ\(\downarrow\) & PAV\(\downarrow\) & AUC\(\uparrow\) & P@0.1\(\uparrow\) & AUC\(\uparrow\) & P@0.1\(\uparrow\) \\ \hline
AniMer-AM
& 66.3 & \textbf{70.7} & 88.1 & 53.8 & 58.8 & 15.5  \\ 
AniMer+
& \textbf{65.6} & 70.9 & \textbf{89.5} & \textbf{59.4} & \textbf{59.3} & \textbf{16.4}   \\
\bottomrule
\end{tabular}%
}
\end{table}

To demonstrate that AniMer+ is more effective in simultaneously handling animals with anatomical differences, we compare the results of AniMer+ training on both mammals and birds with the results of AniMer training on both mammals and birds. We use ViT (AniMer-AM) and ViT-MoE (AniMer+) as the backbones respectively. Note that AniMer-AM and AniMer+ have the same number of \revise{activated} parameters \revise{during a single pass of inference.} The results presented in Table~\ref{tab: Effective of MoE} demonstrate that by incorporating an MoE design, AniMer+ is more effective at simultaneously managing animals with significant gaps in physiological structure.

\subsubsection{\revise{Ablations on the Feature Dimensions of MoE}}
\revise{
We choose a feature dimension of 320 for taxa-specific layer and 960 for taxa-shared layer (1280 in total), resulting in a 1:3 specific-to-shared dimension ratio for MoE.  
We conduct an ablation study in Table~\ref{tab:ablation_moe} to compare the performance of different specific-to-shared dimension ratios within the fixed 1280 full dimension. For 1:1 ratio, both specific and shared dimensions are 640. For 1:4 ratio, the specific dimension is 256 while the shared dimension is 1024. Note that we do not test 1:2 and 1:5 because 1280 can not be divided by 3 nor 6 evenly. From the results tabulated in that table, the 1:3 ratio (320 specific and 960 shared) consistently achieves the best performance across all datasets. 

\begin{table}[ht]
\caption{ \revise{\textbf{Ablation study on feature dimension ratios of taxa-specific and taxa-shared layers.} PAJ: PA-MPJPE. PAV: PA-MPVPE. P@0.1: PCK@0.1.} } 
\label{tab:ablation_moe}
\centering
\resizebox{0.5\textwidth}{!}{
\begin{tabular}{ccccccc}
\toprule
\multirow{2}{*}{Ratio} & \multicolumn{2}{c}{Animal3D}   & \multicolumn{2}{c}{CtrlAni3D} & \multicolumn{2}{c}{Animal Kingdom} \\ \cmidrule(lr){2-3} \cmidrule(lr){4-5} \cmidrule(lr){6-7}
    & PAJ\(\downarrow\)      & PAV\(\downarrow\)      & PAJ\(\downarrow\)      & PAV\(\downarrow\)      & AUC\(\uparrow\)          & P@0.1\(\uparrow\)        \\ \hline
1:1 & 82.8          & 83.0          & 47.1          & 50.0          & 81.0          & 32.8          \\
1:4 & 79.0          & 83.5          & 48.0          & 51.4          & 81.5          & 33.0          \\
1:3 & \textbf{78.0} & \textbf{82.7} & \textbf{44.1} & \textbf{47.6} & \textbf{82.6} & \textbf{33.1} \\ \midrule
\multirow{2}{*}{Ratio} & \multicolumn{2}{c}{CtrlAVES3D} & \multicolumn{2}{c}{CUB}       & \multicolumn{2}{c}{Cow Bird}       \\ \cmidrule(lr){2-3} \cmidrule(lr){4-5} \cmidrule(lr){6-7}
    & PAJ\(\downarrow\)      & PAV\(\downarrow\)      & AUC\(\uparrow\)            & P@0.1\(\uparrow\)        & AUC\(\uparrow\)            & P@0.1\(\uparrow\)       \\ \hline
1:1 & 67.4          & 71.2   & 88.5          & 56.3          & 57.5          & 16.0          \\
1:4 & 66.0          & \textbf{70.9}          & 88.1          & 55.0          & 59.0          & 15.5          \\
1:3 & \textbf{65.6} & \textbf{70.9} & \textbf{89.5} & \textbf{59.4} & \textbf{59.3} & \textbf{16.4} \\ \bottomrule
\end{tabular}
}
\end{table}
}

\subsubsection{\revise{Ablations on Loss Weights}} 
\revise{
The weights for different loss terms are chosen empirically with the general goal of balancing their contributions to the total loss. 
Given the large number of hyperparameters, a comprehensive grid search for all loss weights is not feasible. Therefore, we only perform ablation study for the weight of our proposed animal family supervised contrastive loss \(\mathcal{L}_{con}\), as it is a key component of our method. Experiments are conducted on the Animal Kingdom dataset.
As shown in Table~\ref{tab:ablation_loss_weight}, we test three different values for \(\lambda_{con}\).
\begin{table}[ht]
\caption{\revise{\textbf{Ablation study on the weight of the contrastive loss,} evaluated on the OOD Animal Kingdom dataset.}}
\label{tab:ablation_loss_weight}
\centering
\resizebox{0.35\textwidth}{!}{
\begin{tabular}{cccc}
\toprule
Weight                  & AUC\(\uparrow\)           & PCK@0.1\(\uparrow\)        & PCK@0.15\(\uparrow\)       \\ \hline
0.01                    & 82.6          & 33.8          & 53.4          \\
0.001                   & \textbf{83.0} & \textbf{36.3} & 54.2          \\
0.0005                  & 82.9          & 34.9          & \textbf{54.7} \\ \bottomrule
\end{tabular}
}
\end{table}
While the 0.001 weight achieves the best results on AUC and PCK@0.1 metrics, the 0.0005 weight yields the best performance on the PCK@0.15 metric and is competitive on other metrics. Therefore, for all other experiments, we set $\lambda_{con}=0.0005$. 
}

\subsubsection{\revise{Ablations on Two-Stage Training}}
\revise{ 
As stated before, we employ two-stage training for mammalian models and single-stage training for avian models. The difference stems from the composition of the datasets. Mammalian datasets are large-scaled and have an imbalanced 2D-to-3D data ratio, while avian datasets are smaller-scaled and more balanced. We conduct ablation studies on the stage choices for both mammalian and avian models in Table~\ref{tab:ablation_training_stage}. 

For the mammalian experiments, the results clearly show that the two-stage training strategy consistently and significantly outperforms the single-stage approach across all evaluation datasets (Animal3D, CtrlAni3D, and Animal Kingdom). This finding holds for both 3D metrics (e.g., PA-MPJPE on Animal3D drops from 85.9 to 80.5) and 2D metrics (e.g., AUC on Animal Kingdom increases from 80.6 to 82.9). These results validate our intuition that a two-stage process is beneficial for handling the large size and the imbalanced 2D-to-3D data ratio in the mammalian datasets.

For the avian experiments, the results demonstrates that single-stage training yields superior performance for birds. On the CUB dataset, for instance, single-stage training achieves a much higher PCK@0.1 (74.8 vs. 58.2) and a better AUC (91.9 vs. 89.4). This supports our decision to use a simpler, single-stage process for the smaller and more balanced avian datasets.

\begin{table}[ht]
\caption{\revise{\textbf{Ablation study on the number of training stages.}  PAJ: PA-MPJPE. PAV: PA-MPVPE. P@0.1: PCK@0.1.} }
\label{tab:ablation_training_stage}
\centering
\resizebox{0.5\textwidth}{!}{
\begin{tabular}{ccccccc}
\toprule
\multirow{2}{*}{\makecell{Mammalian \\ Experiment}} & \multicolumn{2}{c}{Animal3D}   & \multicolumn{2}{c}{CtrlAni3D} & \multicolumn{2}{c}{Animal Kingdom} \\ \cmidrule(lr){2-3} \cmidrule(lr){4-5} \cmidrule(lr){6-7}
& PAJ\(\downarrow\) & PAV\(\downarrow\) & PAJ\(\downarrow\) & PAV\(\downarrow\) & AUC\(\uparrow\)  & P@0.1\(\uparrow\)       \\ \hline
Single-stage & 85.9     & 90.4     & 59.5     & 64.2     & 80.6 & 28.6          \\
Two-stage               & \textbf{80.5}  & \textbf{85.7} & \textbf{44.1} & \textbf{47.6} & \textbf{82.9}    & \textbf{34.9}   \\ \midrule
\multirow{2}{*}{\makecell{Avain \\ Experiment}} & \multicolumn{2}{c}{CtrlAVES3D} & \multicolumn{2}{c}{CUB}       & \multicolumn{2}{c}{Cow Bird}       \\ \cmidrule(lr){2-3} \cmidrule(lr){4-5} \cmidrule(lr){6-7}
& PAJ\(\downarrow\) & PAV\(\downarrow\) & AUC\(\uparrow\)      & P@0.1\(\uparrow\)  & AUC\(\uparrow\)  & P@0.1\(\uparrow\)       \\ \hline
Single-stage               & \textbf{65.5}  & \textbf{90.2} & \textbf{91.9} & \textbf{74.8} & \textbf{65.7}    & 19.4            \\
Two-stage & 71.3     & 90.3     & 89.4     & 58.2     & 63.0 & \textbf{20.1} \\ \bottomrule
\end{tabular}
}
\end{table}
}

\section{More results and analyses}
\begin{figure}[bp]
    \centering
    \includegraphics[width=\columnwidth]{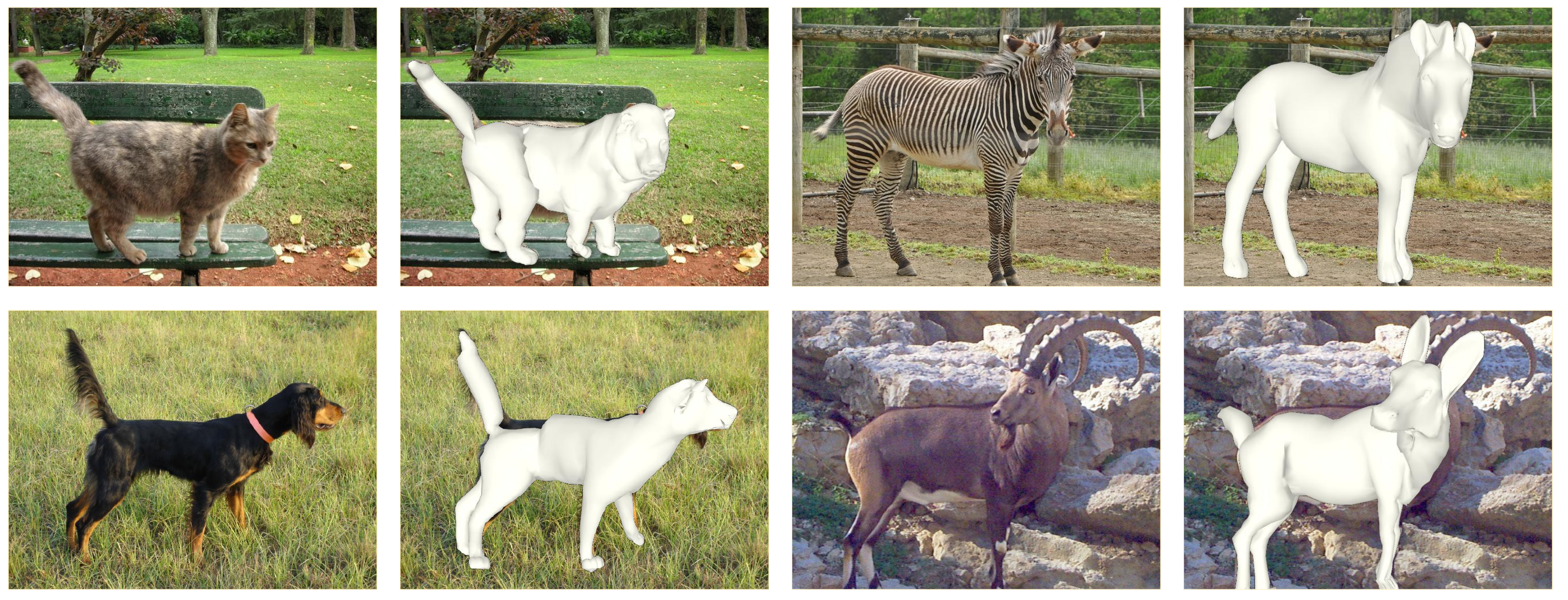}
    \caption{\textbf{Some bad cases in Animal3D.} Overlaid meshes are the ground truth annotations provided by Animal3D. }
    \label{fig:failure cases in Animal3D}
\end{figure}

\noindent\textbf{Comparison Between CtrlAni3D and Animal3D.} CtrlAni3D and Animal3D are both based on SMAL. Both datasets encompass 5 animal families, as presented in Table~\ref{tab:taxonomy}. Animal3D includes more subcategories (e.g., bighorn) compared to CtrlAni3D, which has a greater number of entries. However, 
the Animal3D dataset requires manual 2D annotations for fitting, which introduces error, as shown in the first row of Fig.~\ref{fig:failure cases in Animal3D}. Our CtrlAni3D dataset ensures data quality through cycle consistency and manual filtering.

\noindent\textbf{Comparison Between CtrlAni3D and CG synthetic.} To further emphasize the superiority of our diffusion based gerneration pipeline over traditional CG synthesization, we follow the experiments performed by Animal3D~\cite{xu2023animal3d}, as tabulated in Table~\ref{tab:synthe}. Specifically, ``HMR'' means directly training HMR on Animal3D. ``HMR-Synthetic'' pretrains HMR with a CG-based synthetic dataset generated by textured SMAL~\cite{zuffi2018lions} for 100 epochs before training on Animal3D. Similarly, ``HMR-CtrlAni3D" pretrains HMR on CtrlAni3D for 100 epochs. It is evident that CtrlAni3D provides better pretraining than traditional CG based synthetic data.

\begin{table}[ht]
\centering
\caption{\textbf{CtrlAni3D versus CG synthetic.} The values of ``HMR-Synthetic'' are borrowed from paper ~\cite{xu2023animal3d}.}
\label{tab:synthe}
\resizebox{0.75\columnwidth}{!}{
\begin{tabular}{ccc}
\toprule
& \multicolumn{2}{c}{Animal3D}                                 \\ \cline{2-3} 
\multirow{-2}{*}{Method}          & {PCK@HTH\(\uparrow\)}  & {PA-MPJPE\(\downarrow\)}                      \\ \hline
{HMR}                                &60.5 & 127.8 \\
{HMR-Synthetic}          &63.1 & 124.8 \\
{HMR-CtrlAni3D}  &\textbf{64.0} & \textbf{121.9} 
\\ \bottomrule
\end{tabular}
}
\end{table}

\noindent\textbf{Domain Gap Between CtrlAni3D and Real-world Data.} Although CtrlAni3D can improve model performance on real-world data, there still exists certain domain gap. Table~\ref{tab: domain gap of ctrlani3d dataset} indicates certain domain gap between Animal3D and CtrlAni3D. However, the comparable results between AniMer-M (A3D) and AniMer-M (C3D) on the Animal Kingdom dataset indicate a similar generalizability on in-the-wild data between Animal3D and CtrlAni3D. This is why we aggregate many different datasets for full training. The performance gains of PA-MPJPE and PA-MPVPE shown in Table~\ref{tab: effect of including CtrlAni3D in training} further validate the effectiveness of CtrlAni3D in assisting model generalization.
\begin{table}[htbp]
\caption{\textbf{The generalizability of CtrlAni3D.} A3D means training AniMer-M only on Animal3D, C3D means training AniMer-M only on CtrlAni3D. P@H: PCK@HTH.}
\resizebox{\columnwidth}{!}{%
\label{tab: domain gap of ctrlani3d dataset}
\begin{tabular}{ccccccc}
\toprule
\multirow{2}{*}{Method} & \multicolumn{2}{c}{Animal3D} & \multicolumn{2}{c}{CtrlAni3D} & \multicolumn{2}{c}{Animal Kingdom} \\ \cline{2-7} 
                 & P@H \(\uparrow\) & AUC \(\uparrow\)  & P@H \(\uparrow\) & AUC \(\uparrow\)  & P@H \(\uparrow\) & AUC \(\uparrow\)  \\ \hline
A3D  & 87.0    & 86.0 & 89.7    & 89.9 & 78.0    & 78.6 \\
C3D & 83.8    & 81.9 & 93.5    & 95.0 & 77.8    & 80.3 \\ \bottomrule
\end{tabular}%
}
\end{table}

\noindent\textbf{Comparison Between CtrlAVES3D and CUB17.} Table~\ref{tab: cub17 vs. ctrlani3dplus} shows the results of training solely on CUB17 and training solely on CtrlAVES3D. Our CtrlAVES3D achieves better results than CUB17 plausibly because CtrlAVES3D encompasses 3D annotation and diverse scenes compared to CUB17. 

\begin{table}[ht]
\caption{\textbf{CUB17 versus CtrlAVES3D.} CUB17 means training only on CUB17, CA3D means training only on CtrlAVES3D. P@H: PCK@HTH. }
\resizebox{\columnwidth}{!}{%
\label{tab: cub17 vs. ctrlani3dplus}
\begin{tabular}{ccccccc}
\toprule
\multirow{2}{*}{Method} & \multicolumn{2}{c}{CA3D} & \multicolumn{2}{c}{CUB} & \multicolumn{2}{c}{Cow bird} \\ \cline{2-7} 
                 & P@H \(\uparrow\) & AUC \(\uparrow\)  & P@H \(\uparrow\) & AUC \(\uparrow\)  & P@H \(\uparrow\) & AUC \(\uparrow\)  \\ \hline
CUB17 & 74.1 & 76.3 & 89.3 & 81.0 & 40.9 & 58.7  \\
CA3D  & 96.2 & 92.8 & 90.3 & 81.2 & 51.3 & 62.7 \\ \bottomrule
\end{tabular}%
}
\end{table}

\noindent\textbf{More Discussion About Contrastive Learning.} To further demonstrate the effectiveness of our family-aware contrastive learning scheme, we experiment to compare \(\mathcal{L}_{\text{con}}\) with \(\mathcal{L}_{\text{cls}}\), which simply replaces supervise contrastive loss with cross-entropy loss. Specifically, the contrastive loss directly impacts the feature tokens \( \mathbf{F} \), which in turn indirectly impacts the feature vectors \( \boldsymbol{f} \) and aligns features to model the global structure, capturing family differences. This ensures that the final output shape aligns more closely with the category of the input image. Compared with contrastive learning, \(\mathcal{L}_{\text{cls}}\) (``w $\mathcal{L}_{\text{cls}}$'' in Table~\ref{tab: The impact of the contrastive learning}) focuses solely on optimizing classification accuracy, which may not necessarily improve geometric parameter regression. This is why in Table~\ref{tab: The impact of the contrastive learning} we observe that training with \(\mathcal{L}_{\text{cls}}\) may slightly deteriorate pose and shape estimation. Moreover, contrastive learning facilitates a more compact intra-class distribution and a more separable inter-class distribution in the feature space~\cite{khosla2020supervised}, thereby enhancing the model's capability for few-shot learning. In Table~\ref{tab: The impact of the contrastive learning}, we report the results of AniMer-M and AniMer-A on various specific animal species. It is clearly shown that $\mathcal{L}_{\text{con}}$ can improve performance for animals with limited training samples (e.g., boars constitute less than \(1\%\) of the training set).

\begin{table}[htbp]
\caption{\textbf{Comparison between classifier loss and contrastive loss on real-world datasets (Animal3D, Cow Bird and CUB).} PAJ: PA-MPJPE, PAV: PA-MPVPE, P@0.1: PCK@0.1. }
\resizebox{\columnwidth}{!}{%
\label{tab: The impact of the contrastive learning}
\begin{tabular}{ccccccc}
\toprule
\multirow{2}{*}{Species} & \multicolumn{2}{c}{w \(\mathcal{L}_{\text{cls}}\)} & \multicolumn{2}{c}{w/o \(\mathcal{L}_{\text{con}}\)} & \multicolumn{2}{c}{w \(\mathcal{L}_{\text{con}}\)} \\ \cline{2-7} 
& PAJ\(\downarrow\)  & PAV\(\downarrow\) & PAJ\(\downarrow\) & PAV\(\downarrow\)  & PAJ\(\downarrow\) & PAV\(\downarrow\) \\ \hline
Dog   & 74.9  & 81.3 & 72.1 & 76.7  & \textbf{71.1}  & \textbf{75.1}  \\
Zebra & 66.3  & 68.3 & 60.4 & 63.7  & \textbf{60.6}  & \textbf{62.7}  \\
Horse & 78.5  & 86.6 & 77.4 & 86.5  & \textbf{75.9}  & \textbf{84.1}  \\
Cat   & \textbf{129.4} & \textbf{132.0} & 134.5 & 136.1 & 131.2 & 132.8  \\
Cow   & 83.0  & 86.0 & 80.3 & 84.8  & \textbf{78.1}  & \textbf{83.2}  \\
Sheep & 83.9  & \textbf{88.0} & 83.8 & 91.1  & \textbf{80.1}  & 88.5 \\
Bear  & 79.4  & 80.0 & \textbf{76.5} & 80.5  & 76.8 & \textbf{79.3}  \\
Boar  & 126.5 & 158.7 & 119.1 & 150.5 & \textbf{115.9} & \textbf{142.6} \\ \bottomrule 

\toprule
\multirow{2}{*}{Species} & \multicolumn{2}{c}{w \(\mathcal{L}_{\text{cls}}\)} & \multicolumn{2}{c}{w/o \(\mathcal{L}_{\text{con}}\)} & \multicolumn{2}{c}{w \(\mathcal{L}_{\text{con}}\)} \\ \cline{2-7} 
& AUC\(\uparrow\) & P@0.1\(\uparrow\) & AUC\(\uparrow\) & P@0.1\(\uparrow\)  & AUC\(\uparrow\) & P@0.1\(\uparrow\) \\ \hline

\makecell{Cow \\Bird}  & 60.6 & 14.9 & 61.6 & 15.8 & \textbf{66.3} & \textbf{19.6} \\ \midrule
\makecell{Plieated \\woodpecker} & 88.3 & 51.7 & 88.9 & 56.8 & \textbf{90.2} & \textbf{58.0} \\ \midrule
\makecell{Cedar \\waxwing} & 91.2 & 68.4 & 92.6 & 78.0 & \textbf{92.7} & \textbf{78.4} \\ \midrule
\makecell{Geococcyx} & 88.9 & 50.2 & 91.2 & 65.2 & \textbf{91.4} & \textbf{67.1} \\ \midrule
\makecell{Horned \\puffin} & 90.9 & 63.5 & 92.0 & 78.5 & \textbf{92.4} & \textbf{78.6} \\ \midrule
\makecell{White breasted\\kingfisher} & 90.4 & 57.7 & 92.4 & 75.1 & \textbf{80.1} & \textbf{93.0} \\ \midrule
\makecell{Green \\kingfisher} & 90.3 & 64.8 & 90.0 & 59.4 & \textbf{90.7} & \textbf{66.8} \\ \midrule
\makecell{Blue \\jay} & 90.9 & 62.8 & 91.7 & 68.6 & \textbf{91.8} & \textbf{75.1} \\ \midrule
\makecell{Horned \\puffin} & 90.9 & 63.5 & 92.0 & 78.5 & \textbf{92.4} & \textbf{78.6} \\ \midrule
\makecell{Evening \\grosbeak} & 91.6 & 70.4 & 92.9 & \textbf{79.1} & \textbf{93.5} & 78.6 \\
\bottomrule
\end{tabular}%
}
\end{table}

\section{Failure cases}



\begin{figure}[htbp]
    \centering
    \includegraphics[width=1.0\linewidth]{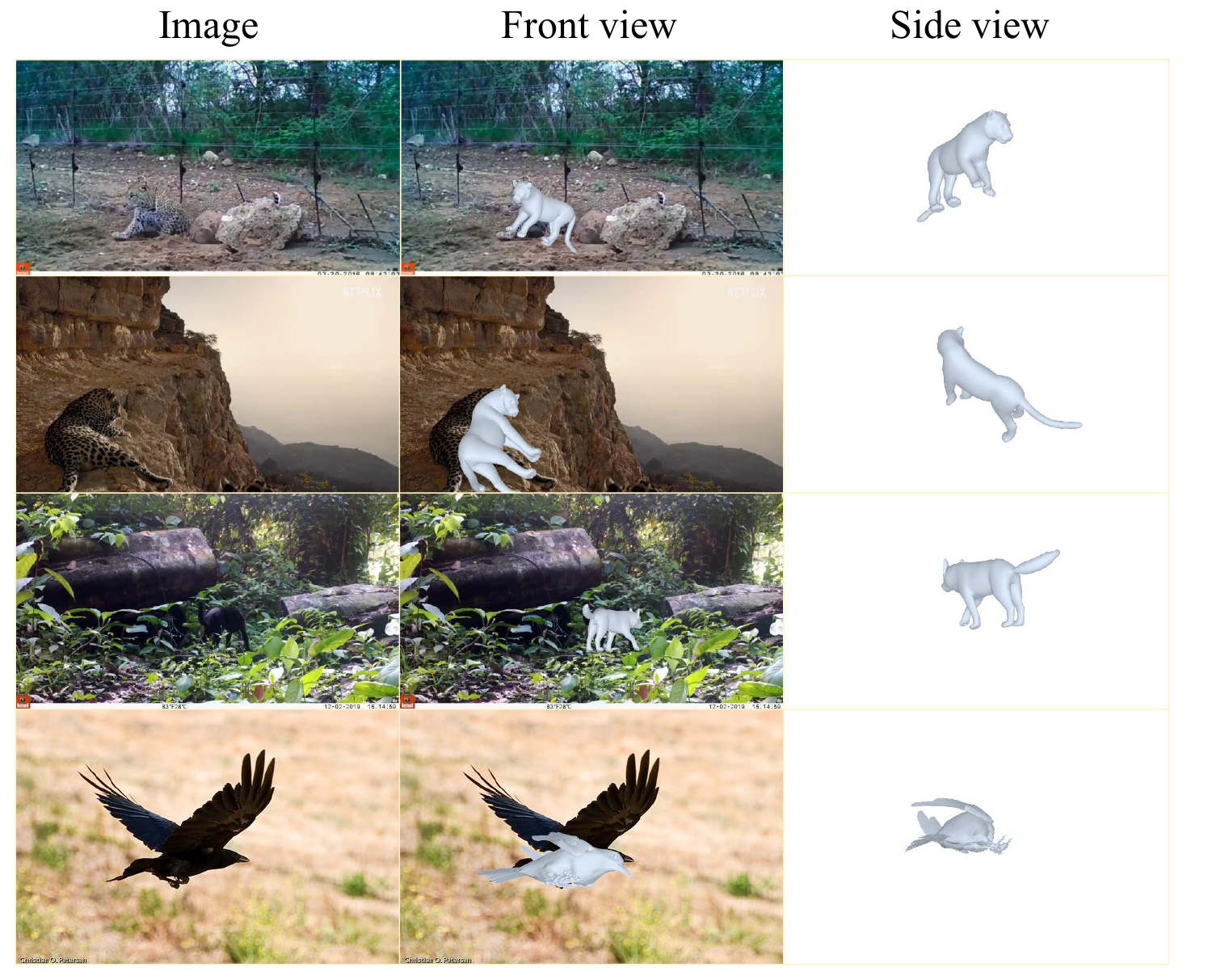}
    \caption{\textbf{Representative failure cases.} }
    \label{fig:failure_cases}
\end{figure}

We provide representative failure cases in Fig.~\ref{fig:failure_cases}. Although our method demonstrates strong robustness, it may fail in certain scenarios. For example, large-scale occlusion (first row), extreme poses (second row), and excessively blurred images (third row) may lead to reconstruction errors. Moreover, our method fails to reconstruct birds in flight because the AVES model is not sufficiently expressive for avian flight poses.



%% file: src/05_conclusion.tex
\section{Conclusion}
\noindent\textbf{Summary. }
In this paper, we first introduce AniMer, a simple yet effective approach for precisely estimating animal pose and shape. Then we extend AniMer to AniMer+, which unifies the mesh recovery of two distinct animal taxa (mammalia and aves) for the first time. The key to the success of AniMer and AniMer+ is a large capacity Transformer backbone together with an aggregated large-scale dataset. For the aggregated dataset, we propose a novel synthetic dataset generation pipeline to generate two novel synthetic general datasets, CtrlAni3D for quadrupeds and CtrlAVES3D for birds, which are rendered by prompting a controllable text-to-image generation model ControlNet. Benefiting from the model design, the synthetic datasets, and our family aware design philosophy, AniMer and AniMer+ outperform previous methods not only on 3D mammalian and avian datasets included for training, but also on out-of-domain 2D real-world mammalian and avian datasets. Moreover, the MoE design also enables AniMer+ to improve the performance across anatomically diverse taxa within a single model. \revise{Our work serves as a blueprint that can readily incorporate new parametric models as they become available.} The principles behind both AniMer and AniMer+ shall inspire the mesh recovery tasks of the whole animal kingdom and enable several downstream applications such as avatar creation and behavioral analysis. 

\noindent\textbf{Limitations and Future Work.}
Despite the demonstrated robustness, our work highlights several avenues for future improvement. Our analysis identifies three main limitations:
(1) {Model Expressivity}: The AVES parametric model, while effective for many bird poses, lacks the expressiveness to accurately represent complex avian articulations, such as those during flight. \revise{For mammalia, the pose space of the SMAL model is based on a limited set of species. For other animal species like primates or rodents, there is a lack of effective parametric models. }
(2) {Challenging Scenarios}: Motion blur and severe occlusions would harm the reconstructed results.
(3) {Data Generation Fidelity}: ControlNet’s species‑specific performance remains uneven, occasionally producing samples that do not correspond to the intended prompt.

In future work, we will address these shortcomings by developing more expressive animal parametric models, \revise{incorporating more varied motion capture data from other species,} adopting spatio‑temporal reconstruction methods, and enhancing ControlNet’s understanding of diverse species. Building upon our current multi-species success, we also aim to expand the model's applicability to an even broader and more diverse range of animals.

\noindent\textbf{Acknowledgements.} 
\revise{This work was done at the Department of Electronic and Electrical Engineering, Southern University of Science and Technology, Shenzhen, China.} 
This study was supported by the National Key Research and Development Program of China (2023YFC2415400); the National Natural Science Foundation of China (T2422012, 62071210, 62125107); the Guangdong Basic and Applied Basic Research (2024B1515020088); the Shenzhen Science and Technology Program (RCYX20210609103056042); the High Level of Special Funds (G030230001, G03034K003); the Shuimu Tsinghua Scholar Program (2024SM324).